\documentclass{article}
\usepackage{amsmath,amsfonts,bm}

\def\eqref#1{equation~\ref{#1}}
\def\1{\bm{1}}

\DeclareMathAlphabet{\mathsfit}{\encodingdefault}{\sfdefault}{m}{sl}
\SetMathAlphabet{\mathsfit}{bold}{\encodingdefault}{\sfdefault}{bx}{n}

\usepackage{fancyhdr}
\usepackage{natbib}
\usepackage{xcolor}
\usepackage{arxiv}

\usepackage{times}
\usepackage[T1]{fontenc}
\usepackage{microtype}
\usepackage{bbm}
\usepackage{graphicx}
\usepackage{caption}
\usepackage{subcaption}
\usepackage{wrapfig}
\usepackage{multirow}
\usepackage{booktabs}
\usepackage[noend]{algpseudocode}
\usepackage{algorithm}
\usepackage{hyperref}
\usepackage{makecell}

\begin{document}

\title{On the Power of Heuristics in Temporal Graphs}

\author{Filip Cornell\thanks{Equal contribution.} \\
Microsoft Gaming \\
\texttt{t-fcornell@microsoft.com} \\
\And
Oleg Smirnov\footnotemark[1] \\
Microsoft Gaming\\
\texttt{oleg.smirnov@microsoft.com} \\
\AND
Gabriela Zarzar Gandler \\
Microsoft Gaming\\
\texttt{gabrielaz@microsoft.com}
\And
Lele Cao\\
Microsoft Gaming\\
\texttt{lelecao@microsoft.com}
}

\maketitle

\begin{abstract}
Dynamic graph datasets often exhibit strong temporal patterns, such as recency, which prioritizes recent interactions, and popularity, which favors frequently occurring nodes. We demonstrate that simple heuristics leveraging only these patterns can perform on par or outperform state-of-the-art neural network models under standard evaluation protocols. To further explore these dynamics, we introduce metrics that quantify the impact of recency and popularity across datasets. Our experiments on BenchTemp~\citep{huang2024benchtemp} and the Temporal Graph Benchmark~\citep{huang2024temporal} show that our approaches achieve state-of-the-art performance across all datasets in the latter and secure top ranks on multiple datasets in the former. These results emphasize the importance of refined evaluation schemes to enable fair comparisons and promote the development of more robust temporal graph models. Additionally, they reveal that current deep learning methods often struggle to capture the key patterns underlying predictions in real-world temporal graphs. For reproducibility, we have made our code publicly available. %\footnote{Redacted for anonymity during review process.}.
\end{abstract}

\section{Introduction}
Dynamic graphs model evolving real-world relationships, where nodes represent entities and edges capture their interactions. These graphs are dynamic, with nodes, edges, weights, or attributes continuously added, removed, or updated over time. Analyzing their temporal patterns is a critical challenge due to their broad applications in fields such as social networks and biological systems. To support this, challenging benchmarks using real-world datasets have been developed, facilitating efficient learning on dynamic graphs~\citep{huang2024temporal, huang2024benchtemp}. A key task in this domain is \textit{link prediction}, which focuses on forecasting future connections between nodes and is foundational for dynamic graph analysis.

Recent methods have increasingly focused on advanced neural network architectures for dynamic graph tasks~\citep{Kumar_2019,Xu2020Inductive,tgn_icml_grl2020,wu2024feasibility,10.5555/3692070.3692716}. However, dynamic graph datasets often exhibit strong recency and popularity patterns that can be effectively captured with simple memorization heuristics. Despite their simplicity, these heuristics have proven to be surprisingly robust baselines, frequently matching or outperforming more complex neural network-based approaches~\citep{poursafaei2022strong,poursafaei2022towards,daniluk2023temporal}.

This work enhances the understanding of recency and popularity in temporal graphs by introducing heuristic algorithms that effectively capture multi-scale temporal patterns. These simple yet powerful methods demonstrate ``unreasonable effectiveness'', outperforming neural models in multiple datasets while also providing a scalable framework for analyzing how temporal dynamics influence ranking behavior.

%, 3) discussing the interplay between temporal patterns, commonly employed sampled evaluation strategies, and their associated limitations, 4) proposing a straightforward method for integrating these heuristics into other models, thereby enhancing their performance, and, finally, 5) introducing a novel dataset specifically designed to mitigate the influence of easily captured patterns, providing a more robust benchmark for evaluating link prediction tasks.

% We argue, much like previous work \cite{poursafaei2022towards}, that a dynamic graph representation should perform better than memorization on any given dataset and therefore outperform the given baselines in this work. We wish for these to serve not only as strong baselines, but as a toolbox in analyzing a dynamic graph dataset to discern the biases contained in them. 

\section{Related Work}
The effect of recency and popularity patterns has been extensively studied in the recommender system literature where it is typically attributed to selection, exposure, presentation, and other biases in interaction data~\citep{chen2023bias,wang2023survey,klimashevskaia2024survey}. Prior research on temporal patterns in dynamic graph datasets has focused on three main directions.

\textbf{Summary metrics.} A range of metrics has been developed to characterize the presence of various temporal patterns. For instance, \citet{poursafaei2022towards} characterized \textit{novelty} (new edges per timestamp), \textit{reoccurrence} (fraction of transductive edges), and \textit{surprise} (test-only edges), demonstrating the challenge of predicting entirely new connections. Similarly, \citet{daniluk2023temporal} proposed statistical distance-based measures to capture both short- and long-term global popularity dynamics, exposing weaknesses in existing evaluation protocols and negative sampling strategies.

\textbf{Tools for interpretation and visualization.} Complementing these metrics are tools designed to make temporal patterns more interpretable. \citet{poursafaei2022towards} introduced Temporal Edge Appearance (TEA) and Temporal Edge Traffic (TET) plots, which reveal when memorization-based approaches may fail -- particularly in sparse graphs or when reoccurrence is low and the surprise index is high. \citet{shirzadkhani2024temporal} later built on this work to provide deeper insights into data characteristics.

\textbf{Leveraging temporal heuristics for prediction.} Beyond measurement and visualization, researchers have proposed models and heuristics to exploit temporal information for prediction tasks. \citet{poursafaei2022towards} presented the EdgeBank heuristic, which achieves strong performance in transductive settings, while \citet{daniluk2023temporal} introduced PopTrack, a simple popularity-based heuristic that outperformed state-of-the-art methods on multiple benchmarks which was then used to create harder negative samples. In a related vein, \citet{poursafaei2022strong} demonstrated that combining structural, interaction-based, and temporal features can produce expressive node representations for accurate classification in both static and dynamic scenarios.% \citet{cong2023we} identified a similar pattern and introduced simple neural network components to account for different aspects of the signal. 

\section{Method}

\subsection{The Notion of Recency}
\label{sec:recency}

Analyses of multiple benchmark datasets indicate that among various link prediction heuristics for dynamic graphs, \textit{recency} (how recently a node has appeared as a destination) emerges as one of the most effective. In many real-world networks, recently active nodes often continue to participate, making recency a robust predictor. Moreover, frequent events also remain highly ranked through continually updated timestamps, reducing the need for added weighting. Building on those observations, the concept of recency is extended to multiple temporal scales, providing a more comprehensive perspective on dynamic graph behavior.

\textbf{Global Recency (GR).} This score identifies the most recently observed destination nodes across the entire graph. Instead of estimating a distribution, a simpler approach records each node's last appearance as a destination node, emphasizing temporal precision through memorization:
${
\text{GR}(v, t) = \max (\{-1\} \cup \{\tau \mid (u, v, \tau) \in \mathcal{G}, \tau < t\}),
% \text{GR}(v, t) = 
% \begin{cases}
% \max \{ \tau \mid (u, v, \tau) \in \mathcal{G}, \tau < t \}, & \text{if such } \tau \text{ exists} \\
% -1, & \text{otherwise},
% \end{cases}
}$
where $\mathcal{G} \subset V \times V \times T$ is the set of temporal edges $(u, v, t)$, and $t \in T$ is a timestamp of the most recent occurrence of node $v \in V$ as a destination.

\textbf{Local Recency (LR).} This score captures the node-level temporal activity of individual destination nodes by focusing on their incoming connections. Rather than relying on a fixed time window, as in EdgeBank, each node retains a time-sorted list of its incoming nodes, effectively reflecting immediate temporal interactions:
${
\text{LR}(u, v, t) = \max (\{-1\} \cup \{\tau \mid (u, v, \tau) \in \mathcal{G}, \tau < t\}),
}$
where $t$ is the timestamp of the most recent interaction between $u$ and $v$.

\subsection{The Notion of Popularity}
\label{sec:pupularity}

As highlighted by \citet{daniluk2023temporal}, many dynamic graph datasets exhibit a pronounced correlation with the historical popularity of destination nodes, reflecting a ``rich-get-richer'' dynamic in which frequently connected nodes continue to attract new links. This effect appears in various real-world systems, where once a node becomes popular, additional edges concentrate around it. Building on this insight, popularity-based heuristics can be implemented analogously to recency-based approaches, capturing how often nodes have served as popular destinations:

\textbf{Global Popularity (GP).} This score counts the total number of times $v$ has appeared as a destination node, being updated at each timestamp:
$
\text{GP}(v, t) = \sum_{\tau < t} \sum_{u' \in V} \mathbbm{1}((u', v, \tau) \in \mathcal{G}),
$
where $\mathbbm{1}(\cdot)$ is the indicator function that equals 1 if the condition holds, and 0 otherwise.

\textbf{Local Popularity (LP).} The score for a node $v$ with respect to a source node $u$ is defined as:
$
\text{LP}(u, v, t) = \sum_{\tau < t} \mathbbm{1}((u, v, \tau) \in \mathcal{G}),
$
where the summation counts the number of times $v$ has appeared as a destination node specifically for source node $u$.

The pseudocode for the proposed heuristics is provided in Algorithm~\ref{alg:heuristics}.

\begin{wrapfigure}[20]{r}{0.44\textwidth}
\begin{minipage}{0.43\textwidth}
\begin{algorithm}[H]
\caption{Recency and Popularity\\ Heuristics}
\label{alg:heuristics}
\begin{algorithmic}[1]
\Require Temporal edges $(u, v, t) \in \mathcal{G}$, metric function $m$
\State Initialize \texttt{LR}, \texttt{GR}, \texttt{LP}, \texttt{GP} as empty dictionaries
\For {$t \in \mathcal{T}$}
    \For {$(u, v) \in \mathcal{G}_t$}
        \For {$h \in \{\texttt{LR}, \texttt{GR}, \texttt{LP}, \texttt{GP}\}$}
            \State Compute $m(h, u, v, t)$
        \EndFor
    \EndFor
    \For {$(u, v) \in \mathcal{G}_t$}
        \State \texttt{LR}[$u$][$v$] $\gets t$
        \State \texttt{GR}[$v$] $\gets t$
        \State \texttt{LP}[$u$][$v$] $\gets$ \texttt{LP}[$u$][$v$] $+ 1$
        \State \texttt{GP}[$v$] $\gets$ \texttt{GP}[$v$] $+ 1$
    \EndFor
\EndFor
\State \Return Scores for \texttt{LR}, \texttt{GR}, \texttt{LP}, \texttt{GP}
\end{algorithmic}
\end{algorithm}
\end{minipage}
\end{wrapfigure}

\subsection{Combining Heuristics}
Tailored heuristics are crucial for different datasets due to their unique characteristics. For example, the Local Recency heuristic performs poorly on the \textit{tgbl-review} dataset because users rarely review the same product twice. This conflicts with the low \textit{novelty} index~\citep{poursafaei2022towards} required for Local Recency, as it struggles to score unseen nodes for a given source. This highlights the need for complementary strategies. Insights from static graph methods, where heuristic combinations leverage individual strengths~\citep{ma2024mixture}, suggest promising directions for extending such approaches to dynamic graphs.

The proposed algorithms also face challenges with ranking ties, which occur when multiple entities receive the same score. Unlike most machine learning models that produce continuous scores, $f \colon \mathcal{G} \rightarrow \mathbb{R}$, these heuristics rely on discrete scoring functions, such as counts or timestamps, i.e., $f \colon \mathcal{G} \rightarrow \mathbb{N}$. For recency-based heuristics, the frequency of ties is influenced by the granularity of dataset timing, with coarse-grained timestamps increasing the likelihood of identical scores. While ties are less common in sampled evaluations with fewer negative examples, they become more prevalent in full-ranking evaluations on datasets with coarser temporal resolution.

An approach in which heuristics are combined addresses this issue by stacking multiple heuristics into a product space, $f \colon \mathcal{G} \rightarrow \mathbb{N}^{|\mathcal{H}|}$, where $\mathcal{H} = \{h_1, h_2, \ldots, h_n\}$ is an ordered set of heuristics. When candidates share the same score under heuristic $h_i$, the next heuristic $h_{i+1}$ determines their internal ranks. This process iterates until ranks are fully resolved, or all heuristics are applied. Structuring the combination this way minimizes ranking ties, reduces discrepancies and improves prediction specificity across datasets. This approach applies to any heuristic in the family $\mathfrak{H} \colon \mathcal{G} \rightarrow S$, where $S \subseteq \mathbb{N}$. For recency heuristics, $S$ represents possible timestamps, while for popularity heuristics, $S = \{0, \ldots, |E|\}$, with $|E|$ as the number of edges. Selecting optimal heuristics for speed and performance depends on the dataset and is left for future study. In this work, unless stated otherwise, we use the order $\text{LR} \rightarrow \text{GR} \rightarrow \text{LP} \rightarrow \text{GP}$.

\section{Experiments and Results}
\label{sec:experiments}
We evaluate the proposed approaches on the TGB~\citep{huang2024temporal} and BenchTemp~\citep{huang2024benchtemp} benchmarks using their respective metrics. As shown in Table~\ref{tab:mrr_tgb}, the heuristic algorithms demonstrate competitive performance, achieving top positions on the TGB leaderboard\footnote{\url{https://tgb.complexdatalab.com/docs/leader_linkprop/}} at the time of writing. Recency mostly outperforms popularity as a predictor across most datasets. However, popularity effectively resolves ties, serving as a complement to methods like LR. Detailed results on both BenchTemp and TGB and comparison to the baseline models are provided in Appendix~\ref{sec:additional_results}.

Two key observations emerge. First, heuristic approaches often outperform modern neural network methods when strong temporal patterns are present. Second, the same dataset, such as \textit{Wikipedia}, can yield different metric values when evaluated under varying protocols, such as those used in TGB and BenchTemp. We hypothesize that these discrepancies stem from two main factors. First, neural network models may struggle to capture dominant temporal patterns, as they are often designed to prioritize long-term dependencies. Second, differences in evaluation protocols can highlight distinct aspects of the data, leading to inconsistencies, especially in sampled settings where results are highly sensitive to the number, quality, and selection of negative examples.

\begin{table}[H]
    \centering
    \scriptsize
% \usepackage{booktabs}

\begin{tabular}{@{}rlllll@{}}
\toprule
\multicolumn{1}{c}{\textbf{Dataset~/~Heuristic}} & \multicolumn{1}{c}{\textbf{LR}} & \multicolumn{1}{c}{\textbf{GR}} & \multicolumn{1}{c}{\textbf{LP}} & \multicolumn{1}{c}{\textbf{GP}} & \multicolumn{1}{c}{\textbf{Combined}} \\ \midrule
\textbf{tgbl-coin}                             & 0.773 (1)                       & 0.613 (3)                       & 0.692 (3)                       & 0.726 (3)                       & 0.809 (1)                             \\
\textbf{tgbl-comment}                          & 0.164 (5)                       & 0.354 (4)                       & 0.106 (7)                       & 0.723 (1)                       & 0.455 (3)                             \\
\textbf{tgbl-flight}                           & 0.831 (1)                       & 0.603 (3)                       & 0.871 (1)                       & 0.183                           & 0.88 (1)                              \\
\textbf{tgbl-review}                           & 0.001                           & 0.321                           & 0.001                           & 0.394     (3)                      & 0.52 (1)                              \\
\textbf{tgbl-wiki}                             & 0.817 (1)                   & 0.04 (12)                       & 0.693 (5)                       & 0.157 (9)                       & 0.821 (1)                             \\ \bottomrule
\end{tabular}
    \caption{Mean Reciprocal Rank (MRR) on TGB~\citep{huang2024temporal} test splits, with leaderboard rankings provided in parentheses.}
    \label{tab:mrr_tgb}
\end{table}

\section{Heuristics as Analysis Tools}
\label{sec:analysis}

\begin{figure}[t]
    \centering
    \begin{subfigure}[t]{0.32\textwidth}
        \centering
        \includegraphics[width=\textwidth]{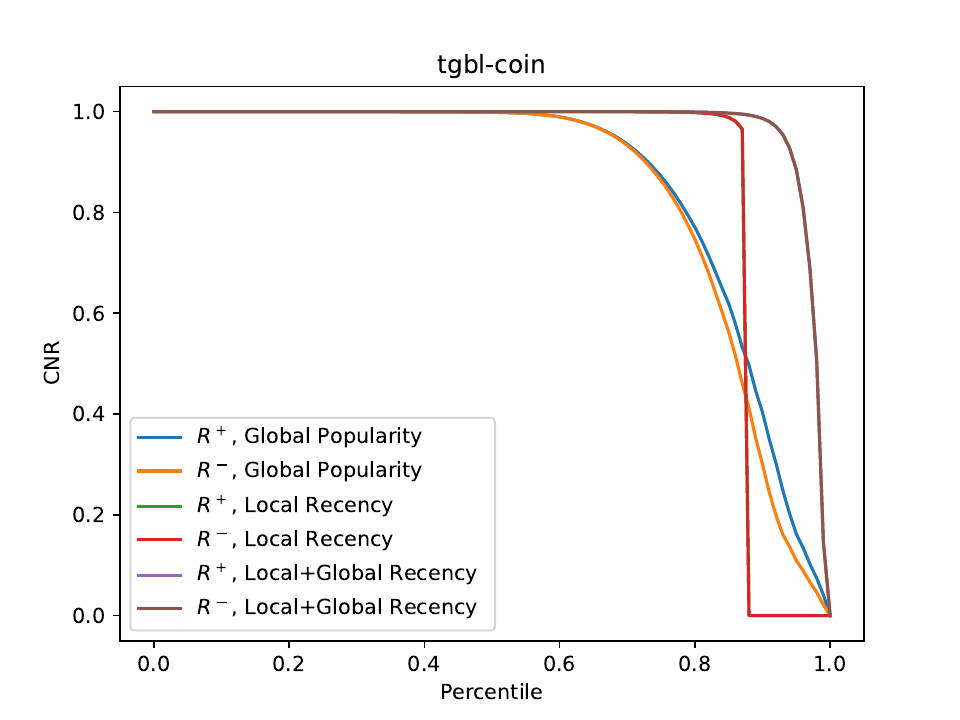}
        \caption{TGB \textit{tgbl-coin} dataset.}
        \label{fig:fpr-1}
    \end{subfigure}
    \hfill
    \begin{subfigure}[t]{0.32\textwidth}
        \centering
        \includegraphics[width=\textwidth]{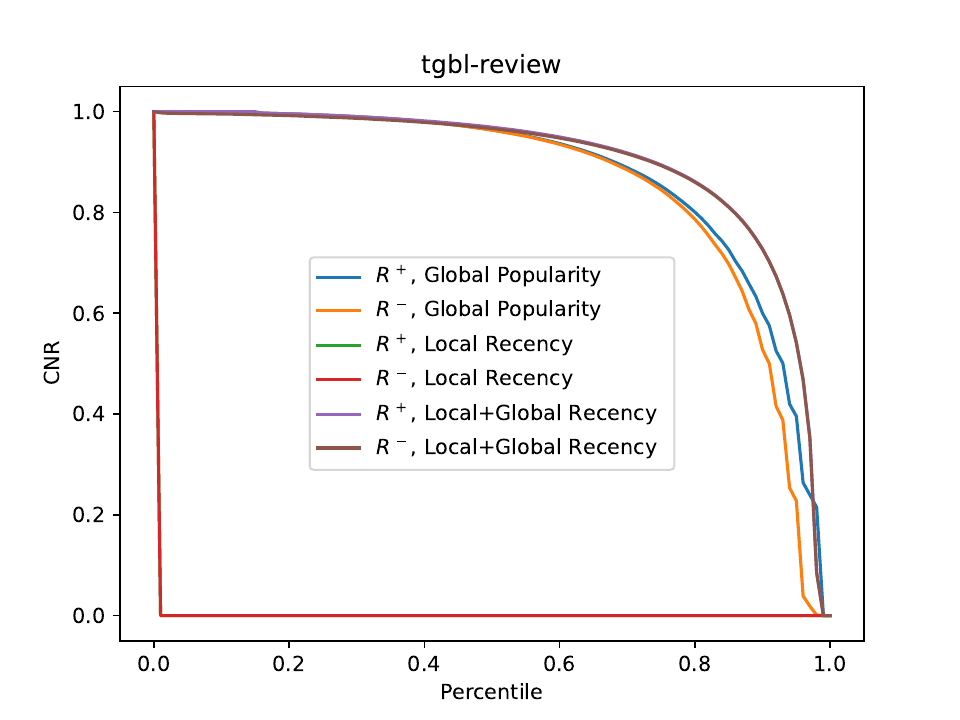}
        \caption{TGB \textit{tgbl-review} dataset.}
        \label{fig:fpr-2}
    \end{subfigure}
    \hfill
    \begin{subfigure}[t]{0.32\textwidth}
        \centering
        \includegraphics[width=\textwidth]{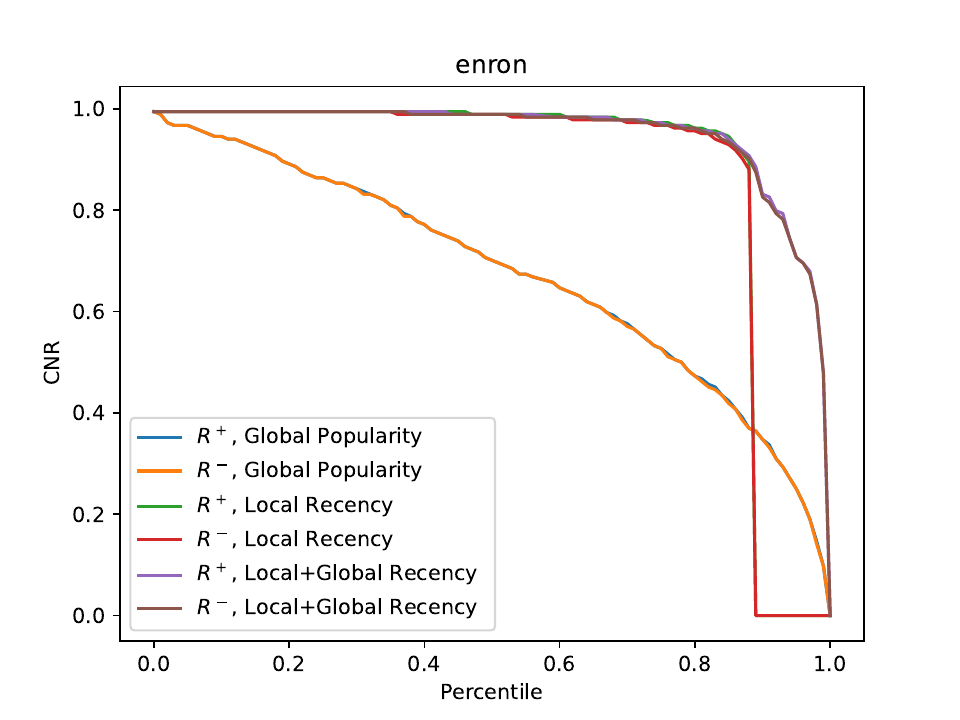}
        \caption{BenchTemp \textit{Enron} dataset.}
        \label{fig:fpr-3}
    \end{subfigure}
    \caption{Complementary Normalized Rank (CNR) plots comparing optimistic ($R^{+}$) and pessimistic ($R^{-}$) ranks across various heuristics and their combinations. Each curve shows a method's performance using the CNR metric across percentiles of all edges, illustrating its ranking effectiveness.}
    \label{fig:full-ranking-plot}
\end{figure}

The prevalence of recency and popularity patterns in a dataset is shaped by its underlying data creation processes. To analyze their impact on ranking behavior, we introduce Complementary Normalized Rank (CNR) metric, computed using \textit{optimistic} ($R^+$) and \textit{pessimistic} ($R^-$) ranks~\citep{ali2021bringing,huang2024temporal}, where $R^+$ assumes favorable tie-breaking, and $R^-$ ranks tied candidates conservatively. At a given $p$, CNR is defined as ${\text{CNR}(p) = 1 - {R_p} / {|E|}}$, indicating that a fraction $p$ of edges was ranked at least as high as $R_p = |E|(1 - \text{CNR}(p))$. While not intended for direct method comparisons, this metric provides insights into dataset predictability and helps assess ranking effectiveness. As shown in Figure~\ref{fig:full-ranking-plot}, CNR plots offer a comprehensive view of how temporal patterns shape ranking performance. While ranking all edges at every timestamp is computationally prohibitive for conventional methods, our heuristics enable efficient computation in logarithmic time with respect to $S$. Further implementation details are provided in Appendix~\ref{sec:implementation_details}, and the discussion of CNR plots in Appendix~\ref{sec:frps}.

\section{Conclusion}
Our findings show that simple and highly efficient heuristics often outperform modern neural network approaches across a range of real-world benchmarks. Their effectiveness depends on dataset characteristics, while the methods introduced provide practical tools for interpreting these patterns and understanding temporal graph behavior.

Inspired by recommender system research, we argue that accurately \textit{modeling} recency and popularity patterns in temporal graph data may not always improve domain-specific metrics, as these patterns often reflect unintended biases such as selection, position, and exposure effects. We defer the exploration of de-biasing techniques for temporal graph datasets, better evaluation protocols, and methods for integrating heuristic signals into neural models to future work.

\subsection*{Acknowledgement}
This work was partially funded by Wallenberg AI, Autonomous Systems and Software Program (WASP).

\bibliography{ref}
\bibliographystyle{iclr2025}

\appendix

\section*{Appendix}

\section{Additional Results}
\label{sec:additional_results}

Table~\ref{tab:full_tgb_leaderboard} presents TGB test MRR results alongside the official leaderboard\footnote{\url{https://tgb.complexdatalab.com/docs/leader_linkprop/}}, showing that our heuristics consistently outperform baseline methods across all cases.  
Table~\ref{tab:full_benchtemp_leaderboard} reports AUC-ROC results for our heuristics compared to the BenchTemp leaderboard\footnote{\url{https://my-website-6gnpiaym0891702b-1257259254.tcloudbaseapp.com/}}, highlighting substantial performance variability across datasets. These results reinforce that each heuristic's effectiveness depends on the presence of the specific temporal pattern it is designed to exploit.

For TGB, we use baseline results from the official leaderboard, except for \textit{tgbl-review} and \textit{tgbl-subreddit}, as the former was updated and the latter was not included at the time of writing. In these cases, we conduct our own evaluations using the reference code, maintaining the TGB hyperparameter configurations. For BenchTemp, we adopt leaderboard baselines and ensure fair comparisons by applying its negative sampling method with a 3-fold evaluation for robust metric estimation.

In the corresponding benchmarks, we compare our approach against JODIE~\citep{kumar2019predicting}, NeurTW~\citep{jin2022neural}, DyGFormer~\citep{yu2023towards}, NAT~\citep{luo2022neighborhood}, TNCN~\citep{zhang2024efficient}, CAWN~\citep{wang2021inductive}, TGN~\citep{rossi2020temporal}, TCL~\citep{wang2021tcl}, TGAT~\citep{Xu2020Inductive}, DyRep~\citep{trivedi2019dyrep}, and GraphMixer~\citep{cong2023we}.

\begin{table}[h]
    \scriptsize
    \centering
    \resizebox{\textwidth}{!}{%
\begin{tabular}{@{}crcccccc@{}}
\toprule
\multicolumn{2}{c}{\textbf{Dataset}}                                          & \textbf{tgbl-wiki} & \textbf{tgbl-coin} & \textbf{tgbl-review} & \textbf{tgbl-comment} & \textbf{tgbl-flight} & \textbf{tgbl-subreddit}                                         \\ \midrule
                                            & DyGFormer                       & \textbf{0.798 $\pm$ 0.004}      & 0.752 $\pm$ 0.004      & ---
                                            % 0.024 $\pm$ 0.005 \tablefootnote{Two runs} 
                                            & \textbf{0.670 $\pm$ 0.001}         & --- & ---
                                            %0.002 $\pm$ 0.000 \tablefootnote{One run - \textcolor{red}{Second run should be done later today}}                                                         
                                            \\
                                            & NAT                             & 0.749 $\pm$ 0.010      & ---                & ---        & ---                   & ---   & ---                                                       \\
                                            & TNCN                            & 0.718 $\pm$ 0.001      & \textbf{0.762 $\pm$ 0.004}      & ---        & \textcolor{blue}{\textbf{0.697 $\pm$ 0.006}}         & 0.820 $\pm$ 0.004 & --- \\
                                            & CAWN                            & 0.711 $\pm$ 0.006      & ---                & ---
                                            % 0.175 $\pm$ 0.000 \tablefootnote{One run}        
                                            & ---                   & ---    & --- %\tablefootnote{One run - \textcolor{red}{The single run should be done later today}}                                                      
                                            \\
                                            & $\text{EdgeBank}_{tw}$                & 0.571              & 0.580              & 0.020                & 0.149                 &0.387   & 0.589      \\
                                            & $\text{EdgeBank}_\infty$ & 0.495              & 0.359              & 0.021                & 0.129                 & 0.167    & 0.485                                                    \\
                                            & TGN                             & 0.396 $\pm$ 0.060      & 0.586 $\pm$ 0.037      & \textcolor{blue}{\textbf{0.414 $\pm$ 0.011}}        & 0.379 $\pm$ 0.021         &  0.705 $\pm$ 0.020 & 0.49 $\pm$ 0.022 \\
                                            & TCL                             & 0.207 $\pm$ 0.025      & ---                & ---        & ---                   & ---  & ---                                                        \\
                                            & TGAT                            & 0.141 $\pm$ 0.007      & ---                & 0.355 $\pm$ 0.012        & ---                   & ---   & 0.388 $\pm$ 0.01                                                       \\
                                            & GraphMixer                      & 0.118 $\pm$ 0.002      & ---                & 0.255 $\pm$ 0.193        & ---                   & ---  & 0.195 $\pm$ 0.001                                                        \\
\multirow{-11}{*}{\textbf{Baseline Models}} & DyRep                           & 0.050 $\pm$ 0.017      & 0.452 $\pm$ 0.046      & 0.106 $\pm$ 0.016        & 0.289 $\pm$ 0.033         &  0.556 $\pm$ 0.014 & 0.113 $\pm$ 0.022 \\ \midrule 
                                            & GR                  & 0.157              & 0.726              & \textbf{0.394}                & \textcolor{red}{\textbf{0.723}}                 & 0.200   & 0.097                                                     \\
                                            & LR                   & \textcolor{blue}{\textbf{0.817}}              & \textcolor{blue}{\textbf{0.773}}              & 0.001                & 0.164                 & \textbf{0.840}  & \textbf{0.716}                                                      \\
                                            & GP               & 0.193              & 0.613              & 0.321                & 0.354                 & 0.619  & 0.193                                                      \\
       & LP                & 0.707              & 0.692              & 0.001                & 0.106                 & \textcolor{blue}{\textbf{0.876}}  & \textcolor{red}{\textbf{0.738}}                                                      \\
\multirow{-5}{*}{\textbf{Heuristics}} & Combined                 & \textcolor{red}{\textbf{0.821}}              & \textcolor{red}{\textbf{0.809}}              & \textcolor{red}{\textbf{0.522}}                & 0.455                 & \textcolor{red}{\textbf{0.88}}  & \textcolor{blue}{\textbf{0.717}}                                                       \\
\bottomrule
\end{tabular}%
}
    \caption{Comparison of the TGB~\citep{huang2024temporal} leaderboard and the proposed heuristics using standardized test MRR, with the \textcolor{red}{\textbf{\textbf{best}}}, \textcolor{blue}{\textbf{second-best}}, and \textbf{third-best} results highlighted in bold and color-coded.}
    \label{tab:full_tgb_leaderboard}
\end{table}

\begin{table}[h]
    \scriptsize
    \centering
    \resizebox{\textwidth}{!}{%
\begin{tabular}{lccccc|ccccccc}
\toprule
\textbf{Dataset} & \multicolumn{5}{c}{\textbf{Heuristics}} & \multicolumn{7}{c}{\textbf{Baseline Models}} \\
\cmidrule(lr){2-6}\cmidrule(lr){7-13}
 & \textbf{Combined} & \textbf{GR} & \textbf{LR} & \textbf{GP} & \textbf{LP} & \textbf{CAWN} & \textbf{DyRep} & \textbf{JODIE} & \textbf{NAT} & \textbf{NeurTW} & \textbf{TGAT} & \textbf{TGN} \\
\midrule
\multicolumn{13}{c}{\textbf{Inductive}}\\
\midrule
\textbf{CanParl} 
& 0.632 $\pm$ 0.002 
& 0.627 $\pm$ 0.002 
& \textbf{0.655 $\pm$ 0.001} 
& 0.630 $\pm$ 0.003 
& 0.653 $\pm$ 0.001 
& \textbf{0.715 $\pm$ 0.097} 
& 0.553 $\pm$ 0.009 
& 0.501 $\pm$ 0.016 
& 0.621 $\pm$ 0.073 
& \textbf{0.887 $\pm$ 0.014} 
& 0.580 $\pm$ 0.007 
& 0.573 $\pm$ 0.027 \\
\textbf{CollegeMsg} 
& \textbf{0.934 $\pm$ 0.000 }
& 0.921 $\pm$ 0.001 
& 0.838 $\pm$ 0.001 
& 0.788 $\pm$ 0.002 
& 0.836 $\pm$ 0.001 
& 0.916 $\pm$ 0.004 
& 0.484 $\pm$ 0.012 
& 0.510 $\pm$ 0.031 
& \textcolor{blue}{\textbf{0.960 $\pm$ 0.017 }}
& \textcolor{blue}{\textbf{0.973 $\pm$ 0.000 }}
& 0.715 $\pm$ 0.001 
& 0.777 $\pm$ 0.052  \\
\textbf{Contact} 
& \textcolor{blue}{\textbf{0.978 $\pm$ 0.000 }}
& 0.875 $\pm$ 0.000 
& 0.966 $\pm$ 0.000 
& 0.623 $\pm$ 0.001 
& 0.939 $\pm$ 0.000 
& 0.969 $\pm$ 0.003 
& 0.865 $\pm$ 0.043 
& 0.936 $\pm$ 0.003 
& 0.947 $\pm$ 0.013 
& \textcolor{blue}{\textbf{0.984 $\pm$ 0.000 }}
& 0.557 $\pm$ 0.005 
& \textbf{0.952 $\pm$ 0.006 } \\
\textbf{Enron} 
& \textcolor{blue}{\textbf{0.940 $\pm$ 0.002 }}
& 0.811 $\pm$ 0.003 
& \textbf{0.918 $\pm$ 0.001 }
& 0.540 $\pm$ 0.004 
& 0.910 $\pm$ 0.003 
& \textbf{0.916 $\pm$ 0.002 }
& 0.712 $\pm$ 0.060 
& 0.804 $\pm$ 0.022 
& \textcolor{red}{\textbf{0.952 $\pm$ 0.006 }}
& 0.905 $\pm$ 0.003 
& 0.553 $\pm$ 0.015 
& 0.816 $\pm$ 0.023  \\
\textbf{Flights} 
& \textbf{0.927 $\pm$ 0.000 }
& 0.784 $\pm$ 0.001 
& 0.851 $\pm$ 0.000 
& 0.827 $\pm$ 0.001 
& 0.850 $\pm$ 0.000 
& \textcolor{red}{\textbf{0.983 $\pm$ 0.000 }}
& 0.869 $\pm$ 0.013 
& 0.922 $\pm$ 0.009 
& \textcolor{red}{\textbf{0.983 $\pm$ 0.003 }}
& 0.916 $\pm$ 0.000 
& 0.832 $\pm$ 0.004 
& \textcolor{blue}{\textbf{0.952 $\pm$ 0.004 }} \\
\textbf{LastFM} 
& \textcolor{blue}{\textbf{0.910 $\pm$ 0.000 }}
& 0.760 $\pm$ 0.001 
& 0.902 $\pm$ 0.000 
& 0.491 $\pm$ 0.000 
& 0.889 $\pm$ 0.000 
& \textbf{0.908 $\pm$ 0.002 }
& 0.799 $\pm$ 0.044 
& 0.801 $\pm$ 0.034 
& \textcolor{red}{\textbf{0.914 $\pm$ 0.004 }}
& 0.884 $\pm$ 0.000 
& 0.520 $\pm$ 0.014 
& 0.828 $\pm$ 0.014  \\
\textbf{MOOC} 
& 0.707 $\pm$ 0.000 
& 0.694 $\pm$ 0.001 
& 0.664 $\pm$ 0.000 
& 0.533 $\pm$ 0.001 
& 0.616 $\pm$ 0.000 
& \textcolor{red}{\textbf{0.948 $\pm$ 0.001 }} 
& \textbf{0.827 $\pm$ 0.018 } 
& 0.778 $\pm$ 0.058 
& 0.733 $\pm$ 0.043 
& 0.804 $\pm$ 0.022 
& 0.737 $\pm$ 0.006 
& \textcolor{blue}{\textbf{0.887 $\pm$ 0.025 }} \\
\textbf{Reddit} 
& \textcolor{red}{\textbf{0.994 $\pm$ 0.008 }}
& 0.904 $\pm$ 0.042 
& 0.933 $\pm$ 0.000 
& 0.799 $\pm$ 0.045 
& 0.933 $\pm$ 0.000 
& \textbf{0.987 $\pm$ 0.000 }
& 0.958 $\pm$ 0.000 
& 0.951 $\pm$ 0.005 
& \textcolor{blue}{\textbf{0.991 $\pm$ 0.003 }}
& 0.980 $\pm$ 0.001 
& 0.965 $\pm$ 0.000 
& 0.976 $\pm$ 0.000  \\
\textbf{SocialEvo} 
& \textcolor{red}{\textbf{0.952 $\pm$ 0.001 }}
& 0.786 $\pm$ 0.000 
& \textcolor{red}{\textbf{0.952 $\pm$ 0.000 }}
& 0.481 $\pm$ 0.002 
& 0.861 $\pm$ 0.000 
& \textcolor{blue}{\textbf{0.930 $\pm$ 0.000 }}
& 0.916 $\pm$ 0.004 
& 0.896 $\pm$ 0.023 
& 0.896 $\pm$ 0.016 
& -- 
& 0.675 $\pm$ 0.002 
& \textbf{0.924 $\pm$ 0.008 } \\
\textbf{Taobao} 
& 0.219 $\pm$ 0.001 
& 0.213 $\pm$ 0.001 
& 0.521 $\pm$ 0.000 
& 0.201 $\pm$ 0.002 
& 0.521 $\pm$ 0.000 
& 0.774 $\pm$ 0.003 
& 0.703 $\pm$ 0.001 
& 0.701 $\pm$ 0.001 
& \textcolor{red}{\textbf{0.999 $\pm$ 0.000 }}
& \textcolor{blue}{\textbf{0.884 $\pm$ 0.002 }}
& 0.526 $\pm$ 0.012 
& \textbf{0.774 $\pm$ 0.003 } \\
\textbf{UCI} 
& \textbf{0.932 $\pm$ 0.001 }
& 0.920 $\pm$ 0.001 
& 0.835 $\pm$ 0.000 
& 0.783 $\pm$ 0.004 
& 0.833 $\pm$ 0.000 
& 0.918 $\pm$ 0.002 
& 0.430 $\pm$ 0.043 
& 0.752 $\pm$ 0.006 
& \textcolor{blue}{\textbf{0.962 $\pm$ 0.017 }}
& \textcolor{red}{\textbf{0.969 $\pm$ 0.003 }}
& 0.702 $\pm$ 0.005 
& 0.808 $\pm$ 0.024 \\
\textbf{UNTrade} 
& 0.632 $\pm$ 0.003
& 0.552 $\pm$ 0.001 
& \textcolor{blue}{\textbf{0.691 $\pm$ 0.001 }}
& 0.473 $\pm$ 0.003 
& 0.594 $\pm$ 0.001 
& \textcolor{red}{\textbf{0.740 $\pm$ 0.001 }}
& 0.647 $\pm$ 0.011 
& 0.673 $\pm$ 0.013 
& 0.648 $\pm$ 0.066 
& 0.592 $\pm$ 0.033 
& -- 
& \textbf{0.673 $\pm$ 0.013 } \\
\textbf{UNVote} 
& 0.54 $\pm$ 0.000 
& 0.517 $\pm$ 0.000 
& \textcolor{blue}{\textbf{0.639 $\pm$ 0.001 }}
& 0.374 $\pm$ 0.001 
& 0.499 $\pm$ 0.001 
& \textbf{0.591 $\pm$ 0.001 }
& 0.499 $\pm$ 0.010 
& 0.512 $\pm$ 0.001 
& \textcolor{red}{\textbf{0.779 $\pm$ 0.008 }}
& 0.586 $\pm$ 0.000 
& 0.477 $\pm$ 0.005 
& 0.588 $\pm$ 0.012  \\
\textbf{USLegis} 
& 0.58 $\pm$ 0.002
& 0.604 $\pm$ 0.002 
& 0.576 $\pm$ 0.004 
& 0.375 $\pm$ 0.003 
& 0.539 $\pm$ 0.004 
& \textcolor{blue}{\textbf{0.967 $\pm$ 0.003 }}
& 0.598 $\pm$ 0.010 
& 0.584 $\pm$ 0.013 
& \textbf{0.745 $\pm$ 0.029 }
& \textcolor{red}{\textbf{0.971 $\pm$ 0.001 }}
& 0.557 $\pm$ 0.008 
& 0.613 $\pm$ 0.005 \\
\textbf{Wikipedia} 
& 0.963 $\pm$ 0.001 
& 0.916 $\pm$ 0.000 
& 0.917 $\pm$ 0.000 
& 0.431 $\pm$ 0.001 
& 0.917 $\pm$ 0.000 
& \textbf{0.989 $\pm$ 0.000 } 
& 0.910 $\pm$ 0.003 
& 0.931 $\pm$ 0.002 
& \textcolor{red}{\textbf{0.996 $\pm$ 0.003 }} 
& \textcolor{blue}{\textbf{0.990 $\pm$ 0.000 }} 
& 0.934 $\pm$ 0.003 
& 0.978 $\pm$ 0.001 \\
\midrule
\multicolumn{13}{c}{\textbf{Inductive (New--New)}}\\
\midrule
\textbf{CanParl} 
& 0.566 $\pm$ 0.004 
& 0.564 $\pm$ 0.003 
& \textbf{0.643 $\pm$ 0.004 }
& 0.561 $\pm$ 0.000 
& 0.644 $\pm$ 0.000 
& \textcolor{blue}{\textbf{0.701 $\pm$ 0.124 }}
& 0.443 $\pm$ 0.007 
& 0.435 $\pm$ 0.009 
& 0.569 $\pm$ 0.033 
& \textcolor{red}{\textbf{0.888 $\pm$ 0.005 }}
& 0.596 $\pm$ 0.007 
& 0.563 $\pm$ 0.040  \\
\textbf{CollegeMsg} 
& \textbf{0.931 $\pm$ 0.001 } 
& 0.920 $\pm$ 0.003 
& 0.843 $\pm$ 0.001 
& 0.729 $\pm$ 0.003 
& 0.842 $\pm$ 0.000 
& 0.930 $\pm$ 0.002 
& 0.527 $\pm$ 0.005 
& 0.532 $\pm$ 0.027 
& \textcolor{blue}{\textbf{0.940 $\pm$ 0.037 }}
& \textcolor{red}{\textbf{0.976 $\pm$ 0.001 }} 
& 0.783 $\pm$ 0.003 
& 0.797 $\pm$ 0.011 \\
\textbf{Contact} 
& \textcolor{blue}{\textbf{0.979 $\pm$ 0.001 }} 
& 0.872 $\pm$ 0.003 
& \textbf{0.968 $\pm$ 0.001 } 
& 0.460 $\pm$ 0.004 
& 0.952 $\pm$ 0.002 
& 0.965 $\pm$ 0.001 
& 0.660 $\pm$ 0.040 
& 0.753 $\pm$ 0.006 
& 0.949 $\pm$ 0.003 
& \textcolor{red}{\textbf{0.982 $\pm$ 0.000 }} 
& 0.545 $\pm$ 0.006 
& 0.912 $\pm$ 0.005 \\
\textbf{Enron} 
& \textcolor{red}{\textbf{0.966 $\pm$ 0.001 }}
& 0.773 $\pm$ 0.004 
& \textbf{0.952 $\pm$ 0.004 }
& 0.404 $\pm$ 0.009 
& 0.953 $\pm$ 0.001 
& \textcolor{blue}{\textbf{0.961 $\pm$ 0.005 }} 
& 0.657 $\pm$ 0.052 
& 0.680 $\pm$ 0.002 
& \textcolor{red}{\textbf{0.969 $\pm$ 0.005 }} 
& 0.939 $\pm$ 0.000 
& 0.531 $\pm$ 0.018 
& 0.764 $\pm$ 0.018 \\
\textbf{Flights} 
& 0.918 $\pm$ 0.001 
& 0.771 $\pm$ 0.003 
& 0.876 $\pm$ 0.000 
& 0.717 $\pm$ 0.003 
& 0.876 $\pm$ 0.000 
& \textcolor{blue}{\textbf{0.987 $\pm$ 0.001 }} 
& 0.890 $\pm$ 0.027 
& 0.930 $\pm$ 0.008 
& \textcolor{red}{\textbf{0.991 $\pm$ 0.001 }} 
& 0.941 $\pm$ 0.000 
& 0.857 $\pm$ 0.006 
& \textbf{0.965 $\pm$ 0.002 } \\
\textbf{LastFM} 
& \textbf{0.966 $\pm$ 0.000 } 
& 0.910 $\pm$ 0.000 
& 0.962 $\pm$ 0.000 
& 0.500 $\pm$ 0.001 
& 0.962 $\pm$ 0.001 
& \textcolor{blue}{\textbf{0.970 $\pm$ 0.000 }} 
& 0.868 $\pm$ 0.016 
& 0.885 $\pm$ 0.009 
& \textcolor{red}{\textbf{0.974 $\pm$ 0.002 }} 
& 0.963 $\pm$ 0.000 
& 0.509 $\pm$ 0.033 
& 0.875 $\pm$ 0.001 \\
\textbf{MOOC} 
& 0.688 $\pm$ 0.001 
& 0.669 $\pm$ 0.003 
& 0.647 $\pm$ 0.003 
& 0.314 $\pm$ 0.009 
& 0.613 $\pm$ 0.000 
& \textcolor{red}{\textbf{0.942 $\pm$ 0.000 }} 
& 0.722 $\pm$ 0.018 
& 0.707 $\pm$ 0.016 
& 0.656 $\pm$ 0.029 
& \textbf{0.805 $\pm$ 0.009 } 
& 0.740 $\pm$ 0.006 
& \textcolor{blue}{\textbf{0.876 $\pm$ 0.004 }} \\
\textbf{Reddit} 
& \textcolor{red}{\textbf{1.000 $\pm$ 0.000 }} 
& 0.938 $\pm$ 0.088 
& 0.875 $\pm$ 0.000 
& 0.979 $\pm$ 0.030 
& 0.875 $\pm$ 0.000 
& \textcolor{blue}{\textbf{0.995 $\pm$ 0.002 }} 
& 0.953 $\pm$ 0.005 
& 0.938 $\pm$ 0.009 
& \textcolor{blue}{\textbf{0.995 $\pm$ 0.001 }} 
& \textbf{0.988 $\pm$ 0.000 }
& 0.960 $\pm$ 0.004 
& 0.981 $\pm$ 0.000 \\
\textbf{SocialEvo} 
& \textcolor{red}{\textbf{0.950 $\pm$ 0.001 }}
& 0.784 $\pm$ 0.006 
& \textcolor{red}{\textbf{0.951 $\pm$ 0.001 }}
& 0.237 $\pm$ 0.004 
& 0.901 $\pm$ 0.001 
& \textcolor{blue}{\textbf{0.932 $\pm$ 0.000 }}
& 0.774 $\pm$ 0.022 
& 0.648 $\pm$ 0.049 
& \textbf{0.928 $\pm$ 0.047 }
& -- 
& 0.466 $\pm$ 0.007 
& 0.879 $\pm$ 0.004 \\
\textbf{Taobao} 
& 0.170 $\pm$ 0.002 
& 0.164 $\pm$ 0.001 
& 0.522 $\pm$ 0.000 
& 0.133 $\pm$ 0.001 
& 0.522 $\pm$ 0.000 
& \textbf{0.785 $\pm$ 0.015 }
& 0.717 $\pm$ 0.001 
& 0.717 $\pm$ 0.001 
& \textcolor{red}{\textbf{1.000 $\pm$ 0.000 }} 
& \textcolor{blue}{\textbf{0.908 $\pm$ 0.001 }} 
& 0.523 $\pm$ 0.005 
& 0.708 $\pm$ 0.001 \\
\textbf{UCI} 
& \textbf{0.929 $\pm$ 0.001 } 
& 0.917 $\pm$ 0.002 
& 0.840 $\pm$ 0.001 
& 0.724 $\pm$ 0.003 
& 0.838 $\pm$ 0.000 
& 0.924 $\pm$ 0.003 
& 0.477 $\pm$ 0.010 
& 0.639 $\pm$ 0.016 
& \textcolor{blue}{\textbf{0.947 $\pm$ 0.026 }} 
& \textcolor{red}{\textbf{0.972 $\pm$ 0.002 }} 
& 0.768 $\pm$ 0.004 
& 0.805 $\pm$ 0.021 \\
\textbf{UNTrade} 
& 0.587 $\pm$ 0.018
& 0.554 $\pm$ 0.003 
& \textcolor{blue}{\textbf{0.704 $\pm$ 0.006 }}
& 0.262 $\pm$ 0.014 
& 0.703 $\pm$ 0.010 
& \textcolor{red}{\textbf{0.746 $\pm$ 0.008 }}
& 0.536 $\pm$ 0.015 
& 0.592 $\pm$ 0.009 
& \textbf{0.688 $\pm$ 0.018 }
& 0.594 $\pm$ 0.060 
& -- 
& 0.507 $\pm$ 0.006  \\
\textbf{UNVote} 
& 0.379 $\pm$ 0.001 
& 0.516 $\pm$ 0.001 
& \textcolor{blue}{\textbf{0.634 $\pm$ 0.002 }}
& 0.145 $\pm$ 0.004 
& \textbf{0.615 $\pm$ 0.003 }
& 0.578 $\pm$ 0.002 
& 0.473 $\pm$ 0.003 
& 0.491 $\pm$ 0.020 
& \textcolor{red}{\textbf{0.720 $\pm$ 0.075 }}
& 0.567 $\pm$ 0.000 
& 0.500 $\pm$ 0.006 
& \textcolor{blue}{\textbf{0.634 $\pm$ 0.002 }} \\
\textbf{USLegis} 
& 0.442 $\pm$ 0.002
& 0.490 $\pm$ 0.003 
& 0.538 $\pm$ 0.004 
& 0.175 $\pm$ 0.003 
& 0.538 $\pm$ 0.004 
& \textcolor{blue}{\textbf{0.974 $\pm$ 0.006 }}
& 0.564 $\pm$ 0.019 
& 0.539 $\pm$ 0.008 
& 0.890 $\pm$ 0.022 
& \textcolor{red}{\textbf{0.979 $\pm$ 0.000 }}
& 0.532 $\pm$ 0.029 
& \textbf{0.890 $\pm$ 0.022 } \\
\textbf{Wikipedia} 
& 0.976 $\pm$ 0.001 
& 0.929 $\pm$ 0.001 
& 0.940 $\pm$ 0.000 
& 0.352 $\pm$ 0.002 
& 0.940 $\pm$ 0.000 
& \textbf{0.993 $\pm$ 0.001 } 
& 0.926 $\pm$ 0.003 
& 0.935 $\pm$ 0.005 
& \textcolor{red}{\textbf{0.998 $\pm$ 0.001 }} 
& \textcolor{blue}{\textbf{0.996 $\pm$ 0.000 }} 
& 0.958 $\pm$ 0.004 
& 0.986 $\pm$ 0.001 \\
\midrule
\multicolumn{13}{c}{\textbf{Inductive (New--Old)}}\\
\midrule
\textbf{CanParl} 
& \textbf{0.648 $\pm$ 0.002 }
& 0.641 $\pm$ 0.002 
& 0.539 $\pm$ 0.020 
& 0.645 $\pm$ 0.004 
& 0.641 $\pm$ 0.002 
& \textcolor{blue}{\textbf{0.723 $\pm$ 0.085 }}
& 0.507 $\pm$ 0.001 
& 0.508 $\pm$ 0.001 
& \textbf{0.628 $\pm$ 0.081 }
& \textcolor{red}{\textbf{0.885 $\pm$ 0.010 }}
& 0.572 $\pm$ 0.006 
& 0.569 $\pm$ 0.022  \\
\textbf{CollegeMsg} 
& \textbf{0.933 $\pm$ 0.000 }
& 0.921 $\pm$ 0.003 
& 0.481 $\pm$ 0.028 
& 0.807 $\pm$ 0.003 
& 0.921 $\pm$ 0.003 
& 0.917 $\pm$ 0.003 
& 0.517 $\pm$ 0.036 
& 0.832 $\pm$ 0.001 
& \textcolor{red}{\textbf{0.973 $\pm$ 0.019 }}
& \textcolor{blue}{\textbf{0.968 $\pm$ 0.002 }}
& 0.701 $\pm$ 0.005 
& 0.772 $\pm$ 0.037 \\
\textbf{Contact} 
& \textcolor{blue}{\textbf{0.978 $\pm$ 0.000 }}
& 0.876 $\pm$ 0.000 
& 0.857 $\pm$ 0.045 
& 0.632 $\pm$ 0.001 
& 0.876 $\pm$ 0.000 
& \textbf{0.969 $\pm$ 0.003 }
& 0.935 $\pm$ 0.003 
& 0.934 $\pm$ 0.003 
& 0.935 $\pm$ 0.020 
& \textcolor{red}{\textbf{0.984 $\pm$ 0.000 }}
& 0.556 $\pm$ 0.004 
& 0.953 $\pm$ 0.005 \\
\textbf{Enron} 
& \textcolor{blue}{\textbf{0.937 $\pm$ 0.001 }}
& 0.817 $\pm$ 0.002 
& 0.692 $\pm$ 0.065 
& 0.559 $\pm$ 0.005 
& 0.785 $\pm$ 0.013 
& \textbf{0.918 $\pm$ 0.003 }
& 0.786 $\pm$ 0.013 
& 0.904 $\pm$ 0.003 
& \textcolor{red}{\textbf{0.949 $\pm$ 0.008 }}
& 0.901 $\pm$ 0.004 
& 0.559 $\pm$ 0.024 
& 0.810 $\pm$ 0.020 \\
\textbf{Flights} 
& 0.928 $\pm$ 0.000 
& 0.786 $\pm$ 0.001 
& 0.847 $\pm$ 0.000 
& 0.839 $\pm$ 0.000 
& 0.786 $\pm$ 0.001 
& \textcolor{blue}{\textbf{0.983 $\pm$ 0.000 }}
& 0.917 $\pm$ 0.011 
& 0.847 $\pm$ 0.000 
& \textcolor{red}{\textbf{0.986 $\pm$ 0.003 }}
& 0.913 $\pm$ 0.000 
& 0.829 $\pm$ 0.004 
& \textbf{0.950 $\pm$ 0.004 } \\
\textbf{LastFM} 
& \textcolor{blue}{\textbf{0.877 $\pm$ 0.000 }}
& 0.676 $\pm$ 0.001 
& 0.698 $\pm$ 0.036 
& 0.486 $\pm$ 0.001 
& 0.676 $\pm$ 0.001 
& \textbf{0.868 $\pm$ 0.003 }
& 0.730 $\pm$ 0.005 
& 0.846 $\pm$ 0.000 
& \textcolor{red}{\textbf{0.914 $\pm$ 0.001 }}
& 0.831 $\pm$ 0.000 
& 0.519 $\pm$ 0.003 
& 0.763 $\pm$ 0.023  \\
\textbf{MOOC} 
& 0.709 $\pm$ 0.001 
& 0.697 $\pm$ 0.001 
& 0.827 $\pm$ 0.013 
& 0.563 $\pm$ 0.002 
& 0.697 $\pm$ 0.001 
& \textcolor{red}{\textbf{0.949 $\pm$ 0.002 }}
& 0.791 $\pm$ 0.048 
& 0.616 $\pm$ 0.000 
& 0.749 $\pm$ 0.046 
& 0.805 $\pm$ 0.024 
& 0.744 $\pm$ 0.006 
& \textcolor{blue}{\textbf{0.881 $\pm$ 0.033 }} \\
\textbf{Reddit} 
& \textcolor{red}{\textbf{0.994 $\pm$ 0.008 }}
& 0.893 $\pm$ 0.049 
& 0.955 $\pm$ 0.000 
& 0.741 $\pm$ 0.067 
& 0.893 $\pm$ 0.049 
& \textcolor{blue}{\textbf{0.985 $\pm$ 0.000 }}
& 0.949 $\pm$ 0.004 
& 0.949 $\pm$ 0.004 
& \textcolor{red}{\textbf{0.995 $\pm$ 0.002 }}
& \textbf{0.979 $\pm$ 0.002 }
& 0.964 $\pm$ 0.000 
& 0.974 $\pm$ 0.000 \\
\textbf{SocialEvo} 
& \textcolor{red}{\textbf{0.952 $\pm$ 0.001 }}
& 0.786 $\pm$ 0.000 
& \textbf{0.918 $\pm$ 0.005 }
& 0.499 $\pm$ 0.001 
& 0.786 $\pm$ 0.000 
& \textbf{0.916 $\pm$ 0.000 }
& 0.895 $\pm$ 0.030 
& 0.858 $\pm$ 0.001 
& 0.879 $\pm$ 0.032 
& -- 
& 0.684 $\pm$ 0.003 
& \textcolor{blue}{\textbf{0.926 $\pm$ 0.009 }}\\
\textbf{Taobao} 
& 0.276 $\pm$ 0.001 
& 0.270 $\pm$ 0.002 
& 0.699 $\pm$ 0.000 
& 0.280 $\pm$ 0.002 
& 0.270 $\pm$ 0.002 
& 0.757 $\pm$ 0.003 
& 0.699 $\pm$ 0.002 
& 0.521 $\pm$ 0.000 
& \textcolor{red}{\textbf{0.999 $\pm$ 0.000 }}
& \textcolor{blue}{\textbf{0.862 $\pm$ 0.003 }}
& 0.527 $\pm$ 0.024 
& \textbf{0.757 $\pm$ 0.003 } \\
\textbf{UCI} 
& \textbf{0.933 $\pm$ 0.001 }
& 0.921 $\pm$ 0.001 
& 0.426 $\pm$ 0.040 
& 0.803 $\pm$ 0.004 
& 0.921 $\pm$ 0.001 
& 0.918 $\pm$ 0.003 
& 0.714 $\pm$ 0.011 
& 0.830 $\pm$ 0.000 
& \textcolor{red}{\textbf{0.975 $\pm$ 0.016 }}
& \textcolor{blue}{\textbf{0.970 $\pm$ 0.004 }}
& 0.684 $\pm$ 0.008 
& 0.802 $\pm$ 0.027 \\
\textbf{UNTrade} 
& \textbf{0.631 $\pm$ 0.002 }
& 0.552 $\pm$ 0.001 
& \textbf{0.631 $\pm$ 0.014 }
& 0.488 $\pm$ 0.003 
& 0.552 $\pm$ 0.001 
& \textcolor{red}{\textbf{0.741 $\pm$ 0.001 }}
& \textcolor{blue}{\textbf{0.665 $\pm$ 0.011 }}
& 0.584 $\pm$ 0.002 
& 0.581 $\pm$ 0.096 
& -- 
& 0.596 $\pm$ 0.037 
& 0.596 $\pm$ 0.017  \\
\textbf{UNVote} 
& 0.546 $\pm$ 0.000 
& 0.517 $\pm$ 0.000 
& 0.502 $\pm$ 0.020 
& 0.389 $\pm$ 0.000 
& 0.517 $\pm$ 0.000 
& \textcolor{blue}{\textbf{0.593 $\pm$ 0.001 }}
& 0.521 $\pm$ 0.008 
& 0.489 $\pm$ 0.001 
& \textcolor{red}{\textbf{0.779 $\pm$ 0.019 }}
& \textbf{0.588 $\pm$ 0.000 }
& 0.479 $\pm$ 0.003 
& \textbf{0.589 $\pm$ 0.011 }  \\
\textbf{USLegis} 
& 0.621 $\pm$ 0.005 
& \textcolor{blue}{\textbf{0.641 $\pm$ 0.002 }}
& 0.567 $\pm$ 0.010 
& 0.443 $\pm$ 0.004 
& \textcolor{blue}{\textbf{0.641 $\pm$ 0.002 }}
& \textcolor{red}{\textbf{0.967 $\pm$ 0.003 }}
& 0.580 $\pm$ 0.021 
& 0.540 $\pm$ 0.004 
& 0.531 $\pm$ 0.100 
& \textcolor{red}{\textbf{0.968 $\pm$ 0.002 }}
& 0.560 $\pm$ 0.009 
& \textcolor{blue}{\textbf{0.641 $\pm$ 0.002 }} \\
\textbf{Wikipedia} 
& 0.957 $\pm$ 0.001 
& 0.910 $\pm$ 0.001 
& 0.882 $\pm$ 0.003 
& 0.466 $\pm$ 0.002 
& 0.910 $\pm$ 0.001 
& \textcolor{blue}{\textbf{0.989 $\pm$ 0.000 }}
& 0.908 $\pm$ 0.004 
& 0.906 $\pm$ 0.000 
& \textcolor{red}{\textbf{0.996 $\pm$ 0.002 }}
& \textbf{0.988 $\pm$ 0.000 }
& 0.918 $\pm$ 0.002 
& 0.970 $\pm$ 0.001  \\
\midrule
\multicolumn{13}{c}{\textbf{Transductive}}\\
\midrule
\textbf{CanParl} 
& 0.731 $\pm$ 0.000 
& 0.722 $\pm$ 0.001 
& 0.723 $\pm$ 0.000 
& 0.717 $\pm$ 0.003 
& 0.722 $\pm$ 0.001 
& 0.720 $\pm$ 0.091 
& \textcolor{blue}{\textbf{0.794 $\pm$ 0.006 }}
& 0.723 $\pm$ 0.001 
& 0.692 $\pm$ 0.072 
& \textcolor{red}{\textbf{0.892 $\pm$ 0.017 }}
& 0.708 $\pm$ 0.022 
& \textbf{0.758 $\pm$ 0.069 } \\
\textbf{CollegeMsg} 
& \textcolor{blue}{\textbf{0.951 $\pm$ 0.002 }}
& 0.923 $\pm$ 0.001 
& 0.870 $\pm$ 0.000 
& 0.875 $\pm$ 0.001 
& 0.923 $\pm$ 0.001 
& 0.916 $\pm$ 0.004 
& 0.573 $\pm$ 0.069 
& 0.870 $\pm$ 0.000 
& 0.906 $\pm$ 0.012 
& \textcolor{red}{\textbf{0.970 $\pm$ 0.000 }}
& 0.808 $\pm$ 0.003 
& \textbf{0.923 $\pm$ 0.001 } \\
\textbf{Contact} 
& \textcolor{blue}{\textbf{0.982 $\pm$ 0.000 }}
& 0.878 $\pm$ 0.000 
& \textbf{0.976 $\pm$ 0.000 }
& 0.794 $\pm$ 0.000 
& 0.878 $\pm$ 0.000 
& 0.969 $\pm$ 0.003 
& 0.928 $\pm$ 0.021 
& 0.938 $\pm$ 0.007 
& 0.946 $\pm$ 0.021 
& \textcolor{red}{\textbf{0.984 $\pm$ 0.000 }}
& 0.558 $\pm$ 0.009 
& 0.977 $\pm$ 0.003  \\
\textbf{Enron} 
& \textcolor{red}{\textbf{0.929 $\pm$ 0.001 }}
& 0.818 $\pm$ 0.002 
& 0.904 $\pm$ 0.001 
& 0.690 $\pm$ 0.003 
& 0.818 $\pm$ 0.002 
& \textbf{0.916 $\pm$ 0.003 }
& 0.799 $\pm$ 0.036 
& 0.829 $\pm$ 0.015 
& \textcolor{blue}{\textbf{0.921 $\pm$ 0.003 }}
& 0.896 $\pm$ 0.005 
& 0.616 $\pm$ 0.021 
& 0.862 $\pm$ 0.017  \\
\textbf{Flights} 
& \textbf{0.965 $\pm$ 0.000 }
& 0.794 $\pm$ 0.000 
& 0.922 $\pm$ 0.000 
& 0.907 $\pm$ 0.000 
& 0.794 $\pm$ 0.000 
& \textcolor{red}{\textbf{0.986 $\pm$ 0.000 }}
& 0.898 $\pm$ 0.006 
& 0.945 $\pm$ 0.007 
& \textcolor{blue}{\textbf{0.975 $\pm$ 0.006 }}
& 0.930 $\pm$ 0.000 
& 0.902 $\pm$ 0.003 
& \textcolor{blue}{\textbf{0.979 $\pm$ 0.003 }} \\
\textbf{LastFM} 
& \textcolor{red}{\textbf{0.899 $\pm$ 0.000 }}
& 0.620 $\pm$ 0.001 
& \textcolor{blue}{\textbf{0.895 $\pm$ 0.000 }}
& 0.615 $\pm$ 0.000 
& 0.620 $\pm$ 0.001 
& \textbf{0.875 $\pm$ 0.001 }
& 0.679 $\pm$ 0.055 
& 0.677 $\pm$ 0.059 
& 0.854 $\pm$ 0.003 
& 0.839 $\pm$ 0.000 
& 0.509 $\pm$ 0.007 
& 0.774 $\pm$ 0.026  \\
\textbf{MOOC} 
& 0.768 $\pm$ 0.001 
& 0.741 $\pm$ 0.001 
& 0.728 $\pm$ 0.001 
& 0.674 $\pm$ 0.001 
& 0.741 $\pm$ 0.001 
& \textcolor{red}{\textbf{0.946 $\pm$ 0.001 }}
& \textbf{0.824 $\pm$ 0.032 }
& 0.790 $\pm$ 0.021 
& 0.757 $\pm$ 0.031 
& 0.807 $\pm$ 0.019 
& 0.739 $\pm$ 0.006 
& \textcolor{blue}{\textbf{0.900 $\pm$ 0.021 }} \\
\textbf{Reddit} 
& 0.976 $\pm$ 0.000
& 0.881 $\pm$ 0.000 
& 0.958 $\pm$ 0.000 
& 0.898 $\pm$ 0.001 
& 0.881 $\pm$ 0.000 
& \textcolor{red}{\textbf{0.989 $\pm$ 0.000 }}
& 0.980 $\pm$ 0.000
& 0.976 $\pm$ 0.001
& \textbf{0.985 $\pm$ 0.002 }
& 0.984 $\pm$ 0.002
& 0.981 $\pm$ 0.000
& \textcolor{blue}{\textbf{0.987 $\pm$ 0.000 }} \\
\textbf{SocialEvo} 
& \textcolor{red}{\textbf{0.957 $\pm$ 0.000 }}
& 0.783 $\pm$ 0.001 
& \textcolor{red}{\textbf{0.957 $\pm$ 0.000 }}
& 0.724 $\pm$ 0.001 
& 0.783 $\pm$ 0.001 
& \textcolor{blue}{\textbf{0.952 $\pm$ 0.000 }}
& 0.902 $\pm$ 0.003 
& 0.867 $\pm$ 0.023 
& 0.920 $\pm$ 0.006 
& -- 
& 0.785 $\pm$ 0.005 
& \textbf{0.934 $\pm$ 0.000 } \\
\textbf{Taobao} 
& 0.744 $\pm$ 0.004 
& 0.617 $\pm$ 0.008 
& 0.664 $\pm$ 0.000 
& 0.752 $\pm$ 0.006 
& 0.617 $\pm$ 0.008 
& 0.771 $\pm$ 0.003 
& 0.841 $\pm$ 0.001
& 0.840 $\pm$ 0.001
& \textcolor{red}{\textbf{0.894 $\pm$ 0.002 }}
& \textcolor{blue}{\textbf{0.876 $\pm$ 0.001 }}
& 0.540 $\pm$ 0.009 
& \textbf{0.865 $\pm$ 0.001 }  \\
\textbf{UCI} 
& \textcolor{blue}{\textbf{0.954 $\pm$ 0.001 }}
& \textbf{0.926 $\pm$ 0.001 }
& 0.875 $\pm$ 0.000 
& 0.877 $\pm$ 0.003 
& \textbf{0.926 $\pm$ 0.001 }
& 0.919 $\pm$ 0.002 
& 0.509 $\pm$ 0.065 
& 0.879 $\pm$ 0.002 
& 0.908 $\pm$ 0.012 
& \textcolor{red}{\textbf{0.967 $\pm$ 0.003 }}
& 0.800 $\pm$ 0.005 
& 0.888 $\pm$ 0.016  \\
\textbf{UNTrade} 
& \textbf{0.73 $\pm$ 0.001 }
& 0.554 $\pm$ 0.001 
& 0.694 $\pm$ 0.001 
& 0.720 $\pm$ 0.001 
& 0.554 $\pm$ 0.001 
& \textcolor{blue}{\textbf{0.751 $\pm$ 0.001 }}
& 0.638 $\pm$ 0.003 
& 0.679 $\pm$ 0.010 
& \textcolor{red}{\textbf{0.783 $\pm$ 0.047 }}
& -- 
& 0.592 $\pm$ 0.037 
& 0.654 $\pm$ 0.010  \\
\textbf{UNVote} 
& \textbf{0.643 $\pm$ 0.001 }
& 0.519 $\pm$ 0.000 
& 0.647 $\pm$ 0.001 
& \textbf{0.643 $\pm$ 0.001 }
& 0.519 $\pm$ 0.000 
& 0.604 $\pm$ 0.002 
& 0.624 $\pm$ 0.031 
& 0.652 $\pm$ 0.008 
& \textcolor{blue}{\textbf{0.678 $\pm$ 0.041 }}
& 0.587 $\pm$ 0.000 
& 0.513 $\pm$ 0.003 
& \textcolor{red}{\textbf{0.718 $\pm$ 0.011 }} \\
\textbf{USLegis} 
& 0.816 $\pm$ 0.004 
& 0.775 $\pm$ 0.004 
& 0.771 $\pm$ 0.007 
& 0.695 $\pm$ 0.009 
& 0.775 $\pm$ 0.004 
& \textcolor{blue}{\textbf{0.964 $\pm$ 0.004 }}
& 0.743 $\pm$ 0.037 
& \textbf{0.828 $\pm$ 0.002 }
& 0.782 $\pm$ 0.026 
& \textcolor{red}{\textbf{0.972 $\pm$ 0.001 }}
& 0.774 $\pm$ 0.006 
& \textbf{0.828 $\pm$ 0.002 } \\
\textbf{Wikipedia} 
& 0.983 $\pm$ 0.000 
& 0.906 $\pm$ 0.000 
& 0.963 $\pm$ 0.000 
& 0.672 $\pm$ 0.002 
& 0.906 $\pm$ 0.000 
& \textcolor{blue}{\textbf{0.989 $\pm$ 0.000 }}
& 0.943 $\pm$ 0.001 
& 0.951 $\pm$ 0.003 
& 0.979 $\pm$ 0.003 
& \textcolor{red}{\textbf{0.991 $\pm$ 0.000 }}
& 0.951 $\pm$ 0.002 
& \textbf{0.985 $\pm$ 0.000 } \\
\bottomrule
\end{tabular}
}
    \caption{Comparison of the BenchTemp~\citep{huang2024benchtemp} leaderboard and the proposed heuristics, with the \textcolor{red}{\textbf{best}}, \textcolor{blue}{\textbf{second-best}}, and \textbf{third-best} results highlighted in bold and color-coded. We observe that, in some cases, the metric becomes fully saturated, likely due to shortcomings in the sampled evaluation scenario, which tends to overestimate performance by including excessively easy negative examples.}
    \label{tab:full_benchtemp_leaderboard}
\end{table}

\section{Efficient Implementation of Heuristics}
\label{sec:implementation_details}

Computing top-$K$ ranking metrics, such as MRR, across an entire dataset is often computationally expensive. A brute-force approach to determine the exact rank of an entity requires scoring all entities against a query, resulting in a complexity of $\mathcal{O}(|V|)$. To speed up evaluation, many methods sample a smaller set of false (negative) examples and rank the positive item within this subset. However, such methods are biased and inconsistent estimators of true ranking metrics~\citep{krichene2020sampled}. Only AUC-ROC has been proven to provide consistent evaluations, where expected values converge to true performance as the sample size grows.

In contrast, the proposed heuristics enable efficient calculation of full rankings for arbitrary queries in $\mathcal{O}(\log S)$ time by leveraging optimized data structures. When scores are integer-based, each score effectively becomes an index in a consolidated list, grouping all nodes with the same score. This arrangement facilitates direct calculation of exact optimistic and pessimistic ranks by summing nodes in indices below (and, for optimistic ranks, equal to) a particular score. For recency-based heuristics, Fenwick Trees~\citep{fenwick1994new} are used for efficient ranking by storing and retrieving these contiguous sums, which reduces the worst-case complexity of computing full ranks from $\mathcal{O}(|V|)$ to $\mathcal{O}(\log S)$, while the memory usage is bounded by $\mathcal{O}(S + |V|)$ where $S$ represents the number of unique timestamps, and $|V|$ is the number of nodes. 

In essence, the algorithm manages edge updates by dynamically tracking their occurrences across timestamps in both global and local settings. This design achieves a balance between precision and scalability, making it well-suited for large-scale temporal graph data. It should be noted that combining heuristics increases overall complexity, as their independence results in higher computational demands compared to using a single heuristic. % In these cases, the resulting complexity becomes $$\mathcal{O}\Bigg(\max_{\{i \in 1,...,|\mathcal{H}|\}} \Big(\min \{n_{i, ties}, R^-_i\}\Big)\Bigg).$$ As a consequence, heuristics with less ties and higher bias towards the heuristic tend to be more efficient. 

To demonstrate effectiveness, we present the runtime measurements in Table~\ref{tab:runtimes}. For heuristic methods, we report the runtime for a single pass through both the training and evaluation sets. Model runtimes are obtained from the BenchTemp leaderboard, where they are executed on GPUs, whereas heuristic methods are run on a CPU.

\section{Complementary Normalized Ranking Plots}
\label{sec:frps}
\begin{figure}[H]
    \centering
    \begin{subfigure}[t]{0.32\textwidth}
        \centering
        \includegraphics[width=\textwidth]{figs/tgbl-review_percentiles_test.pdf}
        \caption{\textit{tgbl-review} dataset.}
        \label{fig:all-fpr-4}
    \end{subfigure}
    \hfill
    \begin{subfigure}[t]{0.32\textwidth}
        \centering
        \includegraphics[width=\textwidth]{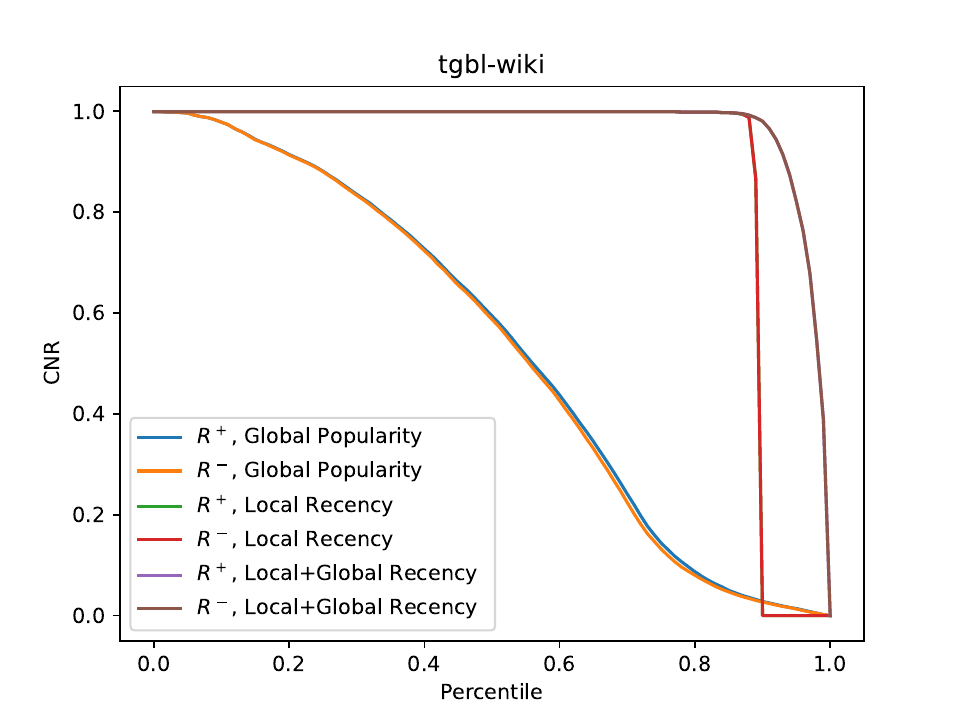}
        \caption{\textit{tgbl-wiki} dataset. }
        \label{fig:all-fpr-1}
    \end{subfigure}
    \hfill
    \begin{subfigure}[t]{0.32\textwidth}
        \centering
        \includegraphics[width=\textwidth]{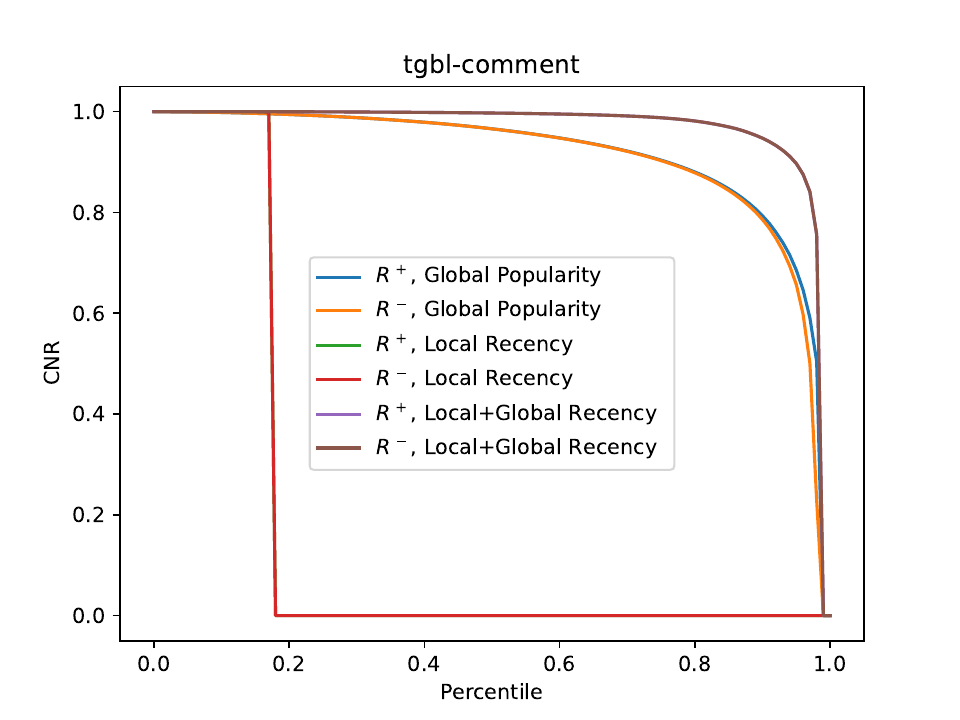}
        \caption{\textit{tgbl-comment} dataset.}
        \label{fig:all-fpr-3}
    \end{subfigure}
    \hfill
    \begin{subfigure}[t]{0.32\textwidth}
        \centering
        \includegraphics[width=\textwidth]{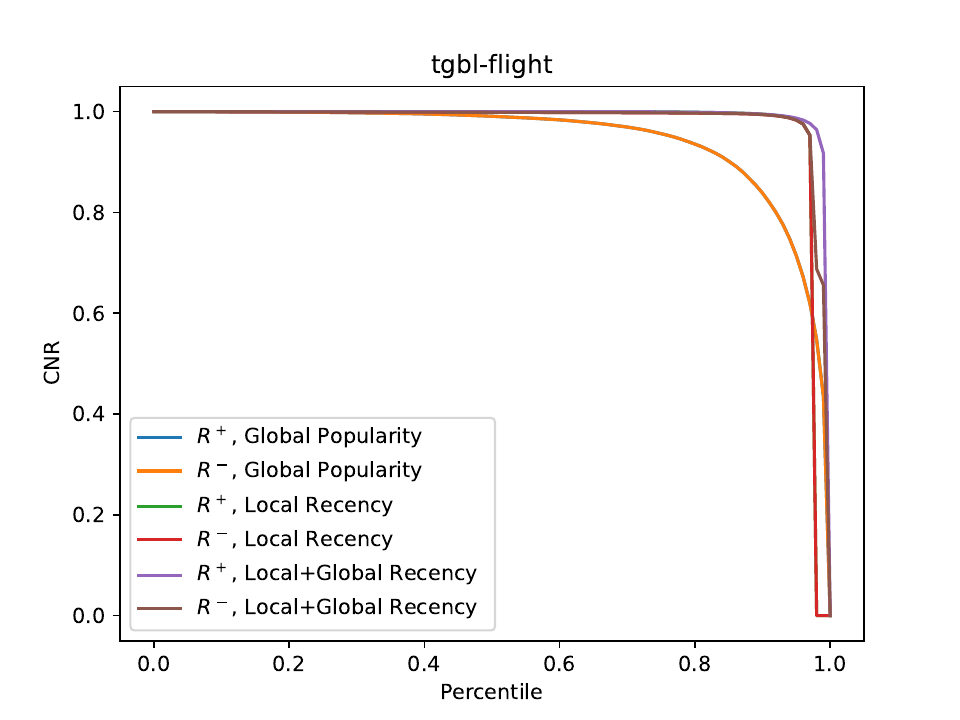}
        \caption{\textit{tgbl-flight} dataset.}
        \label{fig:all-fpr-5}
    \end{subfigure}
    \hfill
    \begin{subfigure}[t]{0.32\textwidth}
        \centering
        \includegraphics[width=\textwidth]{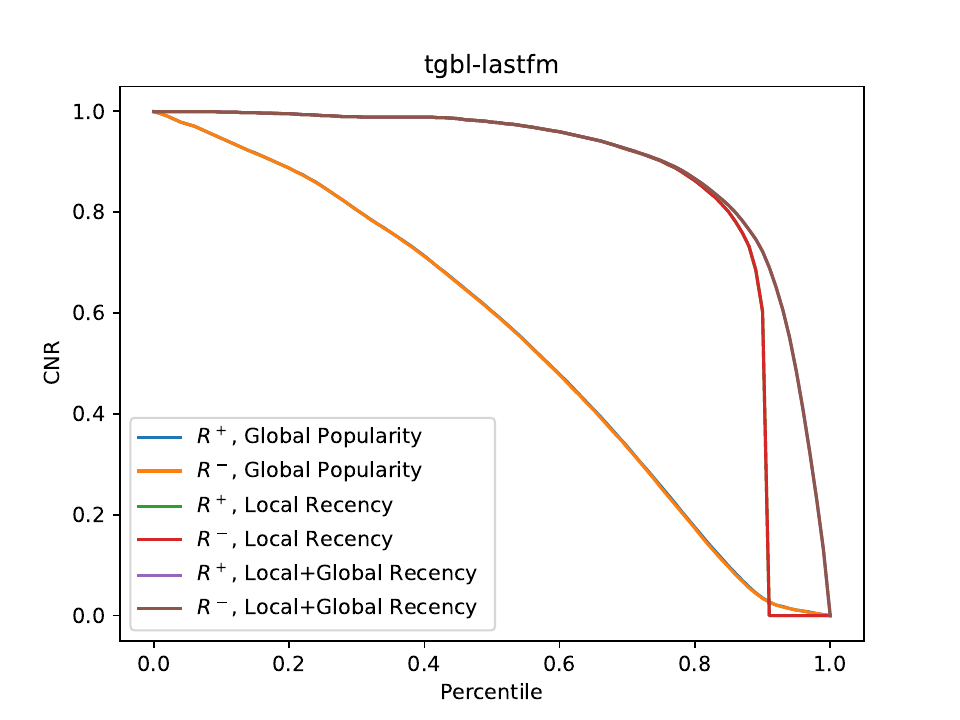}
        \caption{\textit{tgbl-lastfm} dataset.}
        \label{fig:all-fpr-6}
    \end{subfigure}
    \hfill
    \begin{subfigure}[t]{0.32\textwidth}
        \centering
        \includegraphics[width=\textwidth]{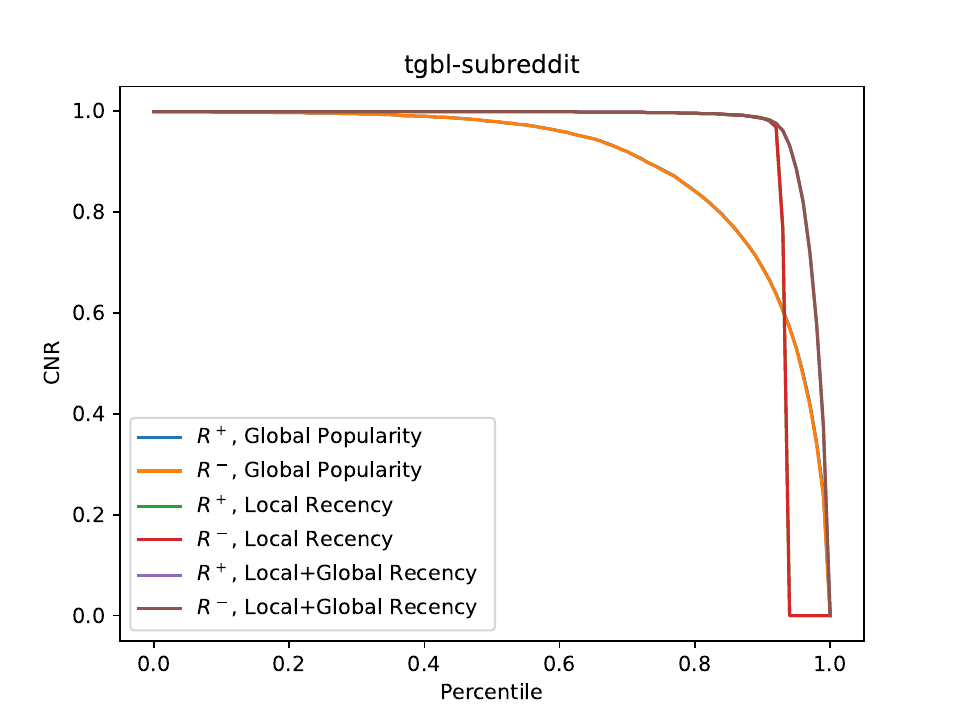}
        \caption{\textit{tgbl-subreddit} dataset.}
        \label{fig:all-fpr-7}
    \end{subfigure}
    \hfill
    \begin{subfigure}[t]{0.32\textwidth}
        \centering
        \includegraphics[width=\textwidth]{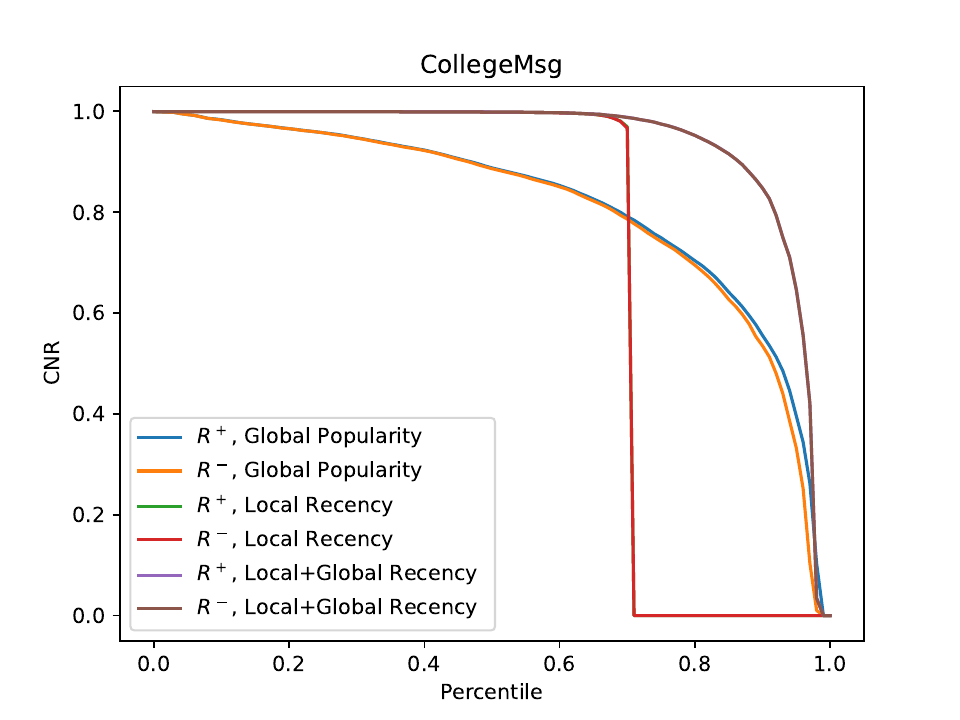}
        \caption{\textit{CollegeMsg} dataset.}
        \label{fig:all-fpr-11}
    \end{subfigure}
    \hfill
    % \begin{subfigure}[t]{0.32\textwidth}
    %     \centering
    %     \includegraphics[width=\textwidth]{figs/Contacts_percentiles_test.pdf}
    %     \caption{\textit{Contacts} dataset.}
    %     \label{fig:all-fpr-12}
    % \end{subfigure}
    % \begin{subfigure}[t]{0.32\textwidth}
    %     \centering
    %     \includegraphics[width=\textwidth]{figs/Flights_percentiles_test.pdf}
    %     \caption{\textit{Flights} dataset.}
    %     \label{fig:all-fpr-14}
    % \end{subfigure}
    % \hfill
    \begin{subfigure}[t]{0.32\textwidth}
        \centering
        \includegraphics[width=\textwidth]{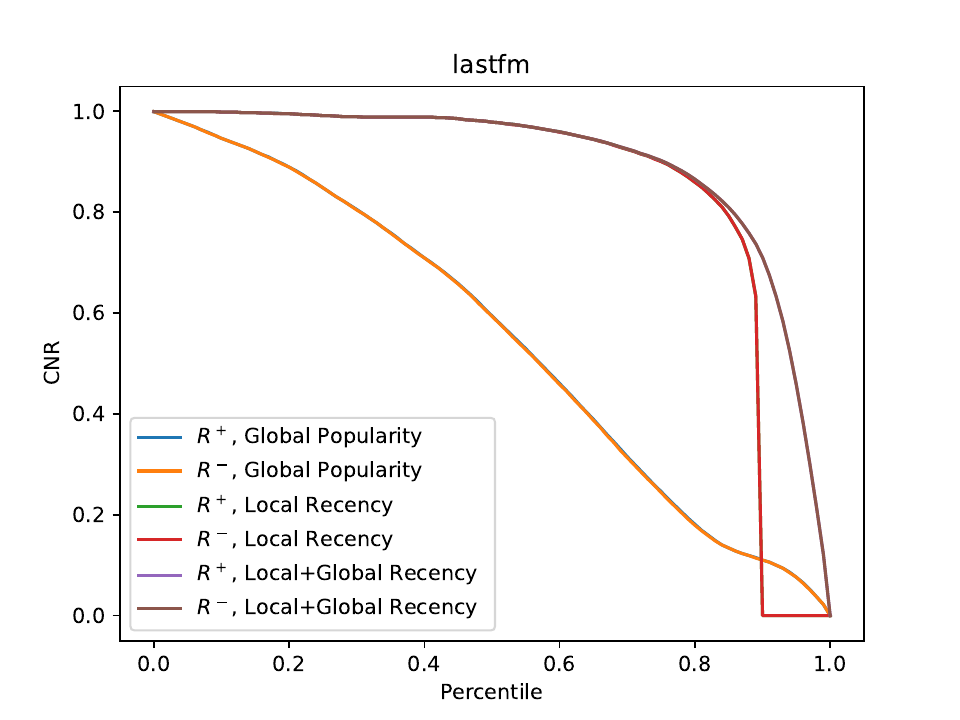}
        \caption{\textit{LastFM} dataset.}
        \label{fig:all-fpr-15}
    \end{subfigure}
    \begin{subfigure}[t]{0.32\textwidth}
        \centering
        \includegraphics[width=\textwidth]{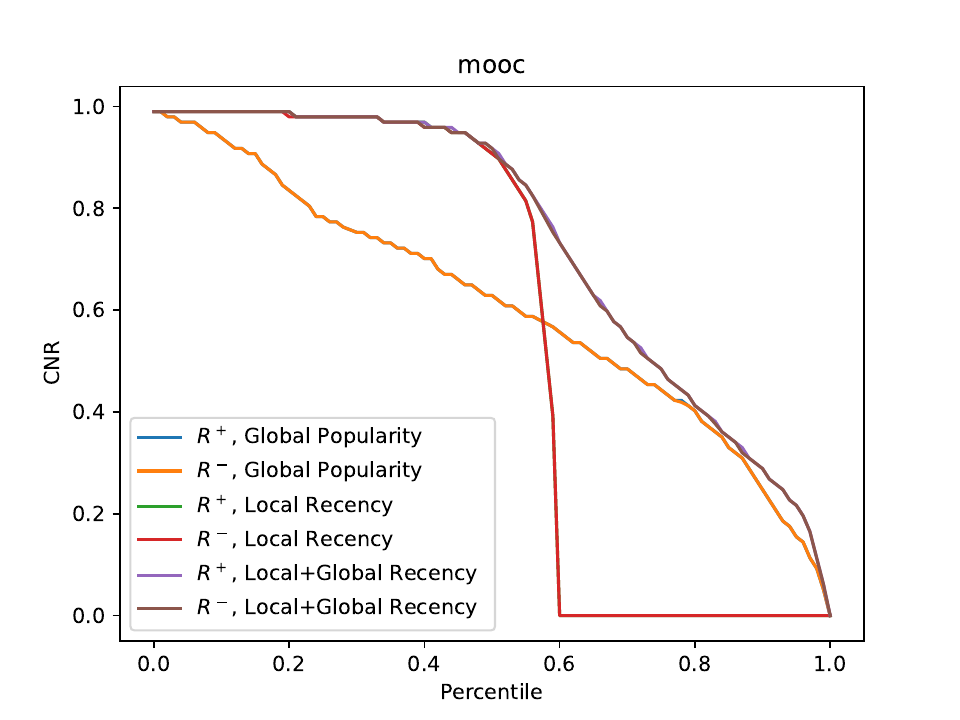}
        \caption{\textit{MOOC} dataset.}
        \label{fig:all-fpr-16}
    \end{subfigure}
    \hfill
    % \begin{subfigure}[t]{0.32\textwidth}
    %     \centering
    %     \includegraphics[width=\textwidth]{figs/reddit_percentiles_test.pdf}
    %     \caption{\textit{Reddit} dataset.}
    %     \label{fig:all-fpr-17}
    % \end{subfigure}
    % \hfill
    \begin{subfigure}[t]{0.32\textwidth}
        \centering
        \includegraphics[width=\textwidth]{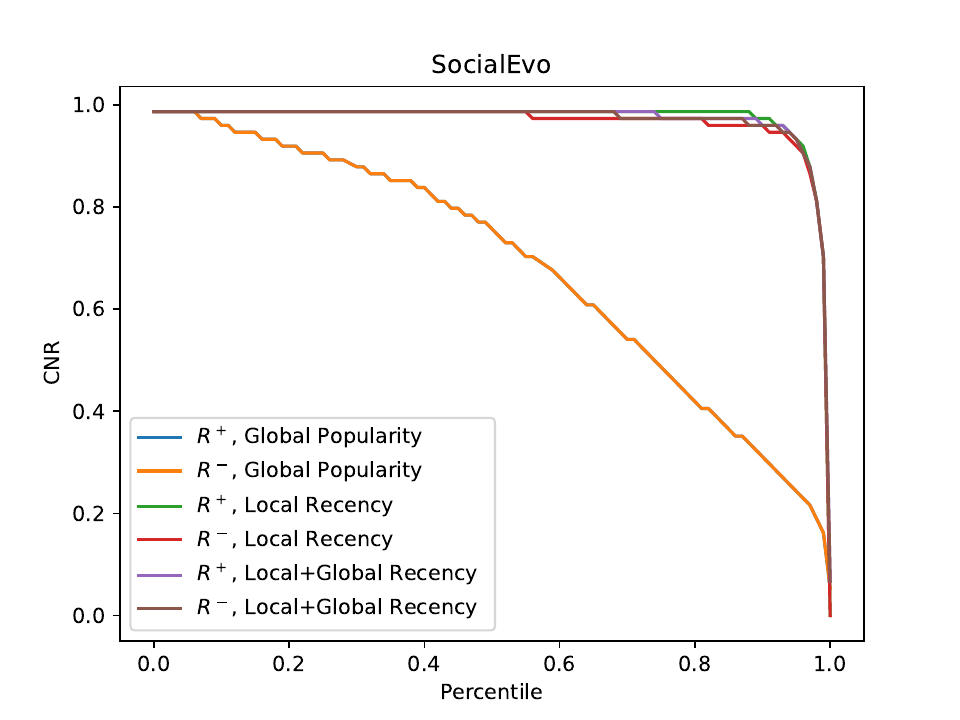}
        \caption{\textit{SocialEvo} dataset.}
        \label{fig:all-fpr-18}
    \end{subfigure}
    \begin{subfigure}[t]{0.32\textwidth}
        \centering
        \includegraphics[width=\textwidth]{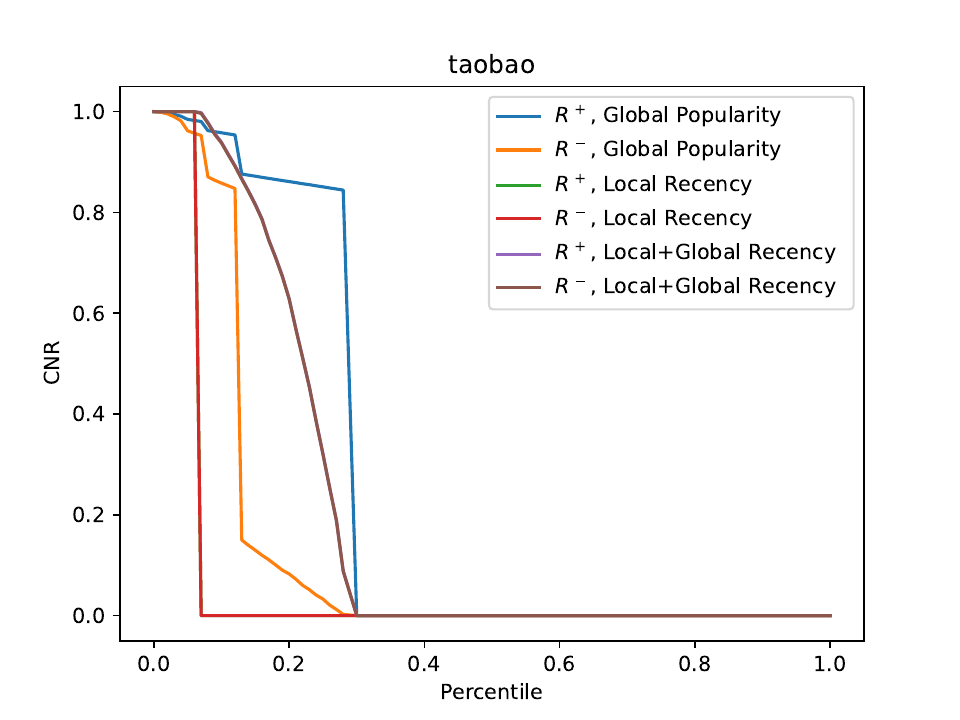}
        \caption{\textit{TaoBao} dataset}
        \label{fig:all-fpr-19}
    \end{subfigure}
    \hfill
    \begin{subfigure}[t]{0.32\textwidth}
        \centering
        \includegraphics[width=\textwidth]{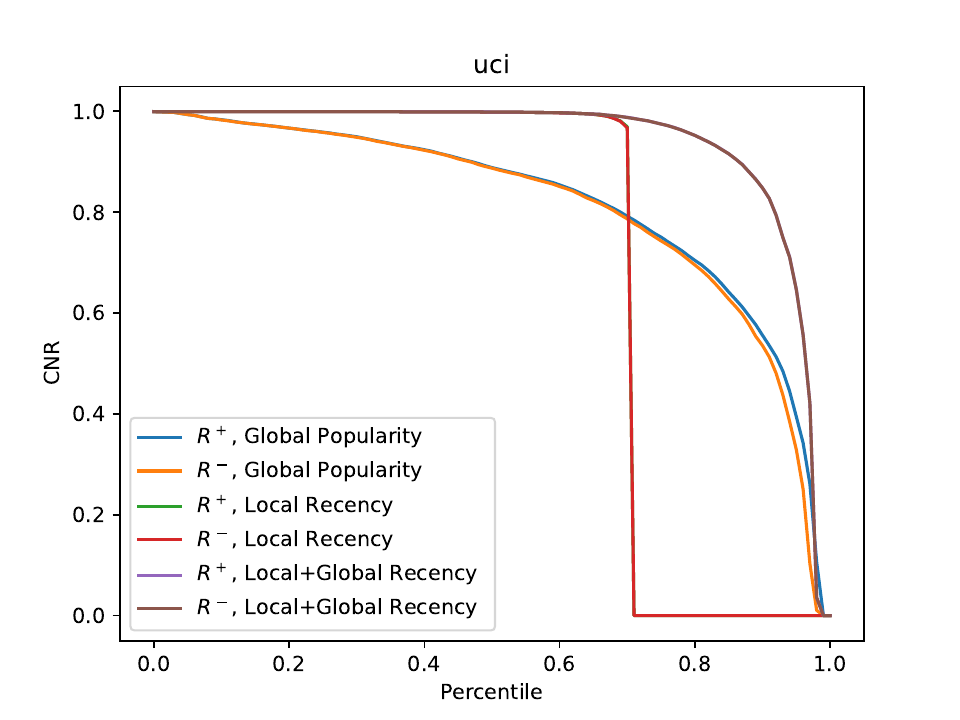}
        \caption{\textit{UCI} dataset.}
        \label{fig:all-fpr-20}
    \end{subfigure}
    \hfill
    \begin{subfigure}[t]{0.32\textwidth}
        \centering
        \includegraphics[width=\textwidth]{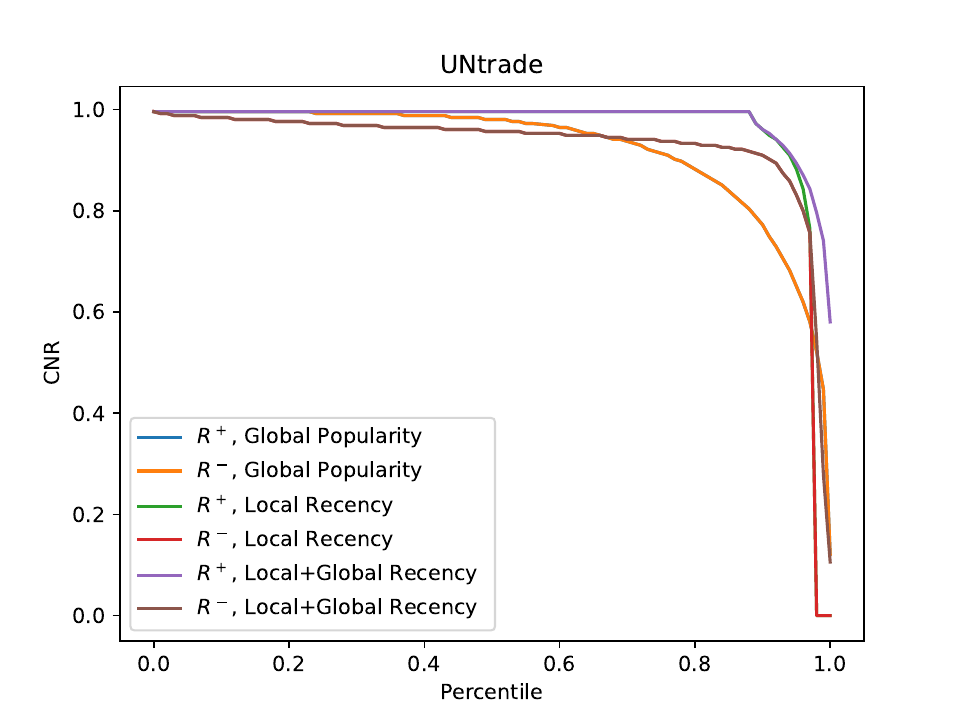}
        \caption{\textit{UNtrade} dataset.}
        \label{fig:all-fpr-21}
    \end{subfigure}
    \begin{subfigure}[t]{0.32\textwidth}
        \centering
        \includegraphics[width=\textwidth]{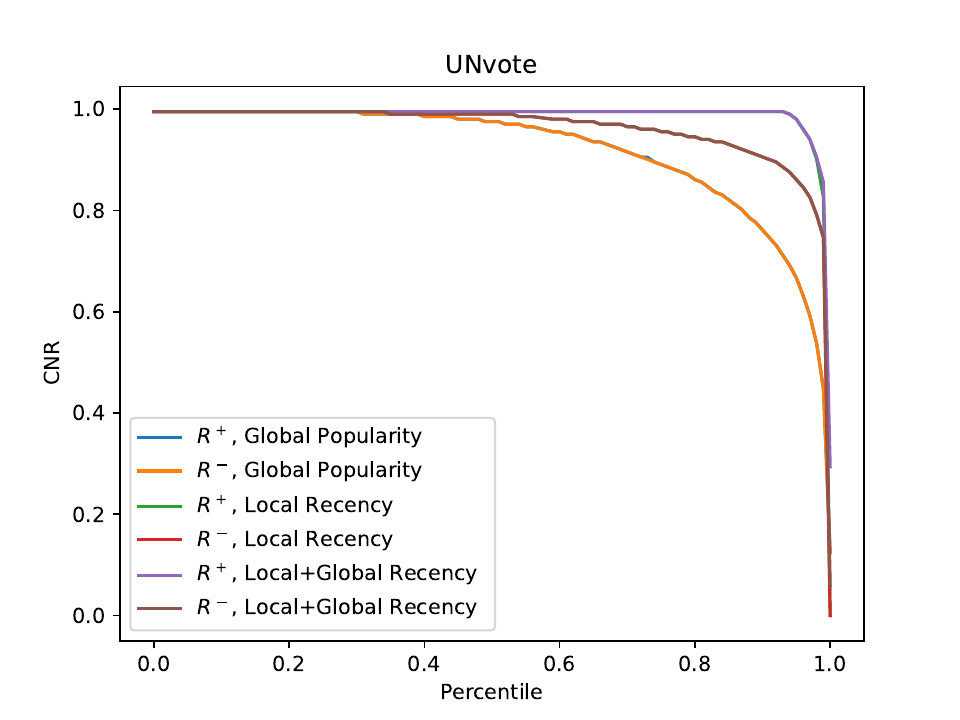}
        \caption{\textit{UNVote} dataset}
        \label{fig:all-fpr-22}
    \end{subfigure}
    \hfill
    \begin{subfigure}[t]{0.32\textwidth}
        \centering
        \includegraphics[width=\textwidth]{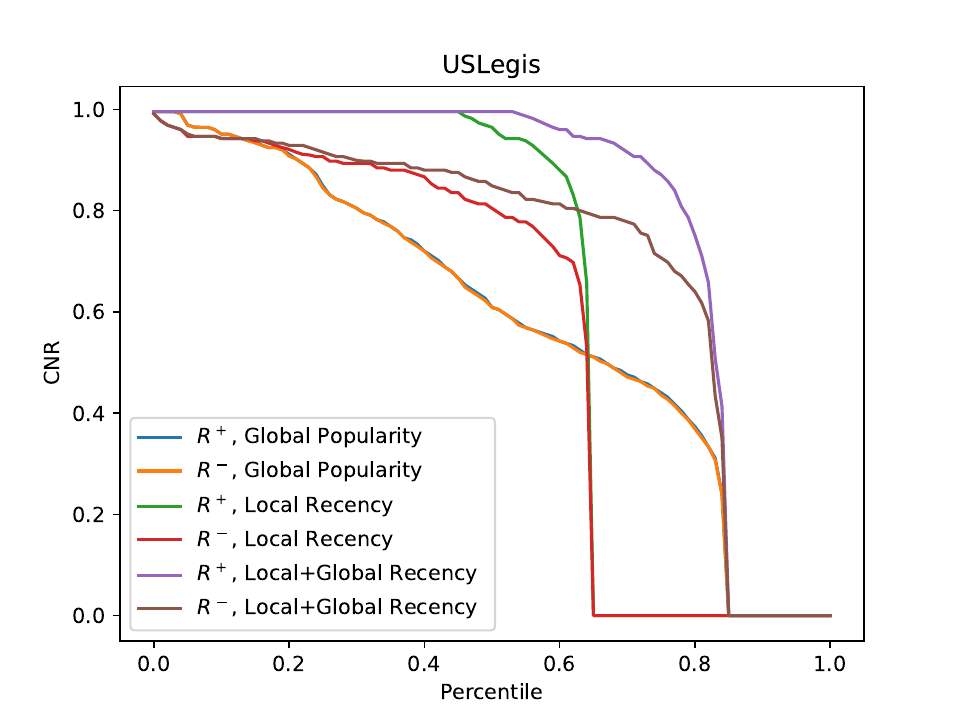}
        \caption{\textit{USLegis} dataset.}
        \label{fig:all-fpr-23}
    \end{subfigure}
    \caption{Complementary Normalized Ranking Plots showing optimistic ($R^{+}$) and pessimistic ($R^{-}$) ranks for different heuristics and their combinations, for TGB (panels a-f) and BenchTemp (panels g-o) datasets. Each curve represents a method's performance across different percentiles of all edges in the dataset, illustrating how well it ranks them overall.}
    \label{fig:full-ranking-plots-1}
\end{figure}

Figure~\ref{fig:full-ranking-plots-1} presents CNR plots for multiple datasets, illustrating how predicted ranks are distributed across different heuristics and highlighting key patterns in how recency and popularity influence ranking performance.

For instance, Figure~\ref{fig:all-fpr-4} reveals the exceptionally poor performance of LR, largely due to approximately 98\% of new edges connecting to previously unseen destination nodes. This aligns with the dataset's nature, where users typically review a product only once, making historical edges unreliable predictors of new interactions. As a result, LR struggles, as it prioritizes recently seen destinations that rarely reappear. To address this, we apply an \textit{inverse} LR heuristic, which penalizes previously seen destinations and prioritizes unseen nodes. This adjustment better reflects the dataset's dynamics, where new interactions are more likely to involve unreviewed products. We further refine this approach by combining inverse LR with GR, significantly improving the resulting performance.

Similarly, Figure~\ref{fig:all-fpr-19} shows minimal recency and popularity effects in the \textit{TaoBao} dataset, a user-item bipartite network with a low average degree of 0.94, consistent with its structural characteristics.

Displaying both pessimistic (\(R^-\)) and optimistic (\(R^+\)) ranks is crucial for evaluating heuristics like Local Recency, which often produce ties. For example, the \textit{USLegis} dataset (Figure~\ref{fig:all-fpr-23}) exhibits a significant gap between \(R^+\) and \(R^-\), reflecting frequent ties. Identifying such discrepancies helps reduce uncertainty and informs whether a single heuristic suffices or if multiple ranking strategies are needed.

%  \begin{algorithm} \caption{Computing Full Rank for Heuristic}
% \label{alg:full_recency} 
% \begin{algorithmic}[1]
% \Require {Temporal edges $(u, v, t) \in \mathcal{G}$, heuristic $\mathcal{H} : \mathcal{G} \rightarrow \mathbb{S}$}
% \State Initialize sparse Fenwick Tree $FT$ with size 
% \State Initialize lists $R^+$ and $R^-$
% \For {$t \in \mathcal{T}$}
%     \For {$(u,v) \in \mathcal{G}_t$}
%         \If{$v$ has been seen for $\mathcal{H}$}
%             \State $R^+_{(u, v, t)} = FT(t) - FT(t_{(u, v)}) + 1$ and append to $R^+$
%             \State $R^-_{(u, v, t)} = FT(t) - FT(t_{(u, v)} - 1) + 1$ and append to $R^-$
%         \Else
%         \EndIf
%     \EndFor
%     \For {$(u,v) \in \mathcal{G}_t$}
%         \State Set $t_{(u, v)} = t$
%     \EndFor
% \EndFor
% \Return Scores for LR and GR
% \end{algorithmic}
% \end{algorithm}
% \begin{table*}[]
%     \centering
%     \include{benchtemp_results}
%     \caption{Benchtemp res table}
%     \label{tab:my_label}
% \end{table*}

\begin{table}[H]
    \centering
    \scriptsize
% \usepackage{booktabs}
\resizebox{\textwidth}{!}{%
\begin{tabular}{@{}r|rrrrrrr|rrrrr@{}}
\toprule
\multicolumn{1}{l|}{}                 & \multicolumn{7}{c|}{\textbf{GPU}}                                                                                                                                                                                                                            & \multicolumn{5}{c}{\textbf{CPU}}                                                                                                                                                 \\ \midrule
\multicolumn{1}{c|}{\textbf{Dataset}} & \multicolumn{1}{c}{\textbf{CAWN}} & \multicolumn{1}{c}{\textbf{DyRep}} & \multicolumn{1}{c}{\textbf{JODIE}} & \multicolumn{1}{c}{\textbf{NAT}} & \multicolumn{1}{c}{\textbf{NeurTW}} & \multicolumn{1}{c}{\textbf{TGAT}} & \multicolumn{1}{c|}{\textbf{TGN}} & \multicolumn{1}{c}{\textbf{LR}} & \multicolumn{1}{c}{\textbf{GR}} & \multicolumn{1}{c}{\textbf{LP}} & \multicolumn{1}{c}{\textbf{GP}} & \multicolumn{1}{c}{\textbf{Combined}} \\ \midrule
\textbf{CanParl}                      & 234.65                            & 2.44                               & 1.96                               & 3.79                             & 4566.24                             & 46.28                             & 3.03                              & 0.3                             & 0.19                            & 0.77                            & 1.64                            & 3.31                                     \\
\textbf{CollegeMsg}                   & 111.96                            & 2.42                               & 1.87                               & 2.83                             & 2097.39                             & 45.8                              & 2.85                              & 0.82                            & 0.93                            & 0.95                            & 1.03                            & 3.65                                     \\
\textbf{Contact}                      & 12100.94                          & 58.20                              & 41.99                              & 96.35                            & 114274.38                           & 1645.1                            & 69.60                             & ---                             & ---                             & ---                             & ---                             & ---                                      \\
\textbf{Enron}                        & 398.38                            & 3.45                               & 2.41                               & 5.70                             & 10896.81                            & 92.24                             & 4.13                              & 0.8                             & 0.94                            & 2.84                            & 3.15                            & 7.45                                     \\
\textbf{Flights}                      & 12105.70                          & 197.61                             & 180.80                             & 76.91                            & 143731.99                           & 1195.3                            & 262.51                            & 17.01                           & 11.42                           & 39.24                           & 49.13                           & 1304.46                                  \\
\textbf{LastFM}                       & 5527.12                           & 39.61                              & 29.42                              & 51.95                            & 51007.3                             & 882.36                            & 45.98                             & 27.42                           & 24.25                           & 25.86                           & 30.64                           & 104.88                                   \\
\textbf{MOOC}                         & 1913.38                           & 33.54                              & 30.19                              & 16.48                            & 13497.27                            & 256.26                            & 41.48                             & 6.16                            & 6.6                             & 6.66                            & 11.21                           & 29.8                                     \\
\textbf{Reddit}                       & 10896.80                          & 4.10                               & 3.40                               & 38.60                            & 5.7                                 & 398.3                             & 92.20                             & 14.39                           & 12.24                           & 18.21                           & 18.91                           & 61.76                                    \\
\textbf{SocialEvo}                    & 12292.31                          & 42.57                              & 27.77                              & 84.72                            & ---                                 & 1544.52                           & 51.35                             & 31.27                           & 27.54                           & 60.78                           & 61.99                           & 176.47                                   \\
\textbf{Taobao}                       & 135.04                            & 38.03                              & 32.17                              & 4.12                             & 1156.96                             & 29.15                             & 34.51                             & 0.73                            & 0.43                            & 1.3                             & 0.68                            & 10.05                                    \\
\textbf{UCI}                          & 121.41                            & 2.49                               & 1.89                               & 2.90                             & 3801.79                             & 57.49                             & 2.72                              & 0.82                            & 0.93                            & 0.95                            & 1.04                            & 3.67                                     \\
\textbf{UNTrade}                      & 1860.84                           & 11.75                              & 7.89                               & 21.57                            & 39402.43                            & ---                               & 14.17                             & 8.25                            & 5.24                            & 15.22                           & 21.16                           & 34.92                                    \\
\textbf{UNVote}                       & 6414.90                           & 25.53                              & 22.31                              & 40.51                            & 88939.57                            & 686.15                            & 28.97                             & 17.57                           & 11.13                           & 32.76                           & 45.64                           & 67.11                                    \\
\textbf{USLegis}                      & 220.41                            & 2.73                               & 2.15                               & 3.16                             & 3208.23                             & 36.69                             & 2.93                              & 0.33                            & 0.27                            & 0.73                            & 1.35                            & 3.21                                     \\
\textbf{Wikipedia}                    & 270.83                            & 10.02                              & 9.37                               & 6.15                             & 4388.59                             & 109.24                            & 12.17                             & 2.48                            & 2.42                            & 3.63                            & 2.63                            & 10.57                                    \\ \bottomrule
\end{tabular}%
}
    \caption{Epoch runtimes for various models, measured in seconds.}
    \label{tab:runtimes}
\end{table}

\end{document}